\documentclass[sigconf,authorversion,nonacm]{acmart}

\AtBeginDocument{%
  }

\usepackage{enumitem}

\usepackage{subfig}

\usepackage{algorithm}
\usepackage{algpseudocode}

\usepackage{amsmath}
\usepackage{mathtools}
\usepackage{amsthm}
\usepackage{xparse}
\usepackage{xspace}

\usepackage{graphicx}
\usepackage{tabularx}
\usepackage{threeparttable}
\usepackage{makecell}
\usepackage{multicol}
\usepackage{multirow}
\usepackage{xparse}
\usepackage{arydshln}

\newcommand{\algo}{\ensuremath{\textsf{FedBiOT}}\xspace}

\newcommand{\mypara}[1]{\paragraph{\textbf{#1}}\xspace}

\newcommand{\param}{\ensuremath{\boldsymbol{w}}\xspace} 
\newcommand{\dinput}{\ensuremath{\boldsymbol{x}}\xspace} 
\newcommand{\dtarget}{\ensuremath{\boldsymbol{y}}\xspace} 
\newcommand{\model}{\ensuremath{\mathcal{M}}\xspace} 
\newcommand{\loss}{\ensuremath{f}\xspace} 
\newcommand{\adapter}{\ensuremath{\mathcal{A}}\xspace} 
\newcommand{\emulator}{\ensuremath{\mathcal{E}}\xspace} 
\newcommand{\extract}{\ensuremath{\textsf{LayerExtract}}\xspace} 

\copyrightyear{2024}
\acmYear{2024}
\setcopyright{rightsretained}
\acmConference[KDD '24]{Proceedings of the 30th ACM SIGKDD Conference on
Knowledge Discovery and Data Mining}{August 25--29, 2024}{Barcelona, Spain}
\acmBooktitle{Proceedings of the 30th ACM SIGKDD Conference on Knowledge
Discovery and Data Mining (KDD '24), August 25--29, 2024, Barcelona,
Spain}\acmDOI{10.1145/3637528.3671897}
\acmISBN{979-8-4007-0490-1/24/08}
\makeatletter
\gdef\@copyrightpermission{
 \begin{minipage}{0.3\columnwidth}
 \href{https://creativecommons.org/licenses/by/4.0/}{\includegraphics[width=0.90\textwidth]{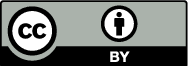}}
 \end{minipage}\hfill
 \begin{minipage}{0.7\columnwidth}
 \href{https://creativecommons.org/licenses/by/4.0/}{This work is licensed under a Creative Commons
Attribution International 4.0 License.}
 \end{minipage}
 \vspace{5pt}
}
\makeatother

\begin{document}

\title{FedBiOT: LLM Local Fine-tuning in Federated Learning without Full Model}

\author{Feijie Wu}
\authornote{Work was done while Feijie Wu was an intern at Alibaba Group.}
\affiliation{%
  \institution{Purdue University}
  \city{}
  \country{}
}
\email{wu1977@purdue.edu}

\author{Zitao Li}
\affiliation{%
  \institution{Alibaba Group}
  \city{}
  \country{}
}
\email{zitao.l@alibaba-inc.com}

\author{Yaliang Li}
\affiliation{%
  \institution{Alibaba Group}
  \city{}
  \country{}
}
\email{yaliang.li@alibaba-inc.com}

\author{Bolin Ding}
\affiliation{%
 \institution{Alibaba Group}
 \city{}
  \country{}
}
\email{bolin.ding@alibaba-inc.com}

\author{Jing Gao}
\affiliation{%
  \institution{Purdue University}
  \city{}
  \country{}
}
\email{jinggao@purdue.edu}

\renewcommand{\shortauthors}{Feijie Wu, Zitao Li, Yaliang Li, Bolin Ding, and Jing Gao}

\begin{abstract}
Large language models (LLMs) show amazing performance on many domain-specific tasks after fine-tuning with some appropriate data. However, many domain-specific data are privately distributed across multiple owners. Thus, this dilemma raises the interest in how to perform LLM fine-tuning in federated learning (FL). However, confronted with limited computation and communication capacities, FL clients struggle to fine-tune an LLM effectively. To this end, we introduce \algo, a resource-efficient LLM fine-tuning approach to FL. Specifically, our method involves the server generating a compressed LLM and aligning its performance with the full model. Subsequently, the clients fine-tune a lightweight yet important part of the compressed model, referred to as an adapter. Notice that as the server has no access to the private data owned by the clients, the data used for alignment by the server has a different distribution from the one used for fine-tuning by clients. We formulate the problem into a bi-level optimization problem to minimize the negative effect of data discrepancy and derive the updating rules for the server and clients. We conduct extensive experiments on LLaMA-2, empirically showing that the adapter has exceptional performance when reintegrated into the global LLM. The results also indicate that the proposed \algo significantly reduces resource consumption compared to existing benchmarks, all while achieving comparable performance levels. 
\end{abstract}



\maketitle

\section{Introduction} 

The recent advancements in large language models (LLMs) have demonstrated incredible performance in various tasks, such as question-answering and problem-solving. This success owes to the pretraining on large datasets, covering a wide range of linguistic patterns and general knowledge. However, in specific domains such as legal advice \cite{nay2023large, cui2023chatlaw} and medical diagnosis \cite{thirunavukarasu2023large, singhal2023large, wang2023chatcad}, LLMs may not provide professional responses because the terminology and context significantly differ from general language use. To address this limitation and enable the generation of domain-specific content, it becomes imperative to fine-tune LLMs with relevant data. This fine-tuning process allows the models to learn from the specific instances and nuances of the target application, ensuring their capability within specialized fields. The quality and quantity of the task-specific data are directly related to the performance of the fine-tuned model on downstream tasks: large and well-labeled data can significantly improve the model, while small and irrelevant data can only benefit the model marginally. However, there are many cases where task-specific data are possessed by multiple data parties, while each of them may have a limited number of samples that can be used to fine-tune LLMs. For example, a hospital in a rural area may only have a limited number of lung cancer cases recorded in its own system; if an LLM is only fine-tuned on one set of those cases, it may not obtain comprehensive knowledge and easily be overfitted. 

To incorporate all the distributed data in the fine-tuning of LLMs, one may consider the batch fine-tuning as follows. If we demand all the data owners to share their data with the LLM server, then LLM fine-tuning could be conducted at the server side. For example, some LLM owners offer fine-tuning APIs as services, but the users must pack their data as files and upload them to use a black-box fine-tuning~\citep{chatgptfinetune}. Apparently, this setup is not applicable to users who have privacy concerns. Especially, some businesses are subject to data privacy regulations~\citep{GDPR2016a, CCPA}, which makes it challenging to share local data with LLM server. 

Therefore, a more practical setting is to let individual data owners keep their data locally, run fine-tuning locally and aggregate the fine-tuning results at the LLM server. This fits well the \emph{federated learning} (FL) framework, which is a distributed paradigm that places a paramount emphasis on preserving privacy. Its conventional algorithms, such as FedAvg \cite{mcmahan2017communication}, are considered practical solutions to overcome data barriers across different data owners. In this paradigm, data owners are treated as clients, and an LLM server coordinates the computation. The standard FL workflow involves three steps repeatedly: (i) The server distributes the global model to all clients; (ii) Each client trains the model locally for multiple iterations and sends the updated model to the server; (iii) The server aggregates the models from the clients and updates the global model accordingly. Despite the potential of this method to facilitate collaborative fine-tuning of an LLM without sharing local data, its feasibility is hindered by two main limitations:
\begin{itemize}[leftmargin=1em]
    \item \textbf{Access to full model of state-of-the-art LLMs:} There exist some open-source LLMs whose model the public can download and have full access to their parameters. However, the most recent and powerful versions of LLMs are usually closed-sourced, i.e, the architecture and parameters are not available to the public. The best closed-source LLMs still have leading performance on a wide range of language tasks, and its leading edge can be maintained or even enhanced after fine-tuning, making it a better choice. As aforementioned, using the blackbox fine-tuning service provided by these closed-source LLMs often violates data users' privacy requirements. Therefore, a federated learning framework that conducts collaborative fine-tuning with the assumption of no access to the full model of LLMs at the client side is more desirable.
    \item \textbf{Computation and communication costs:} Existing federated learning framework could also suffer from the computation and communication challenges when conducting collaborative fine-tuning on LLMs. The fine-tuning process for LLMs entails substantial computational demands and communication costs due to the vast number of trainable model parameters. Clients with limited computational power may struggle to perform complex model updates, leading to prolonged training times or potential disruptions. The transfer of expensive models between the server and the clients also incurs substantial communication costs, leading to substantial bandwidth consumption and increased communication latency. At the server side, there could be network congestion when clients send back their updated huge amount of parameters concurrently.
\end{itemize}

In this paper, we aim to tackle these two challenges and propose to design an effective and practical collaborative LLM fine-tuning framework. To address the first challenge, We follow the setting proposed in offsite-tuning~\citep{xiao2023offsite} and its federated version FedOT~\citep{kuang2023federatedscope}. We assume that the LLM owner does not collect data directly from clients but serves as the server in FL, who can use a public dataset to distill her LLM and aggregate some local updates on part of the model from clients; multiple clients want to collaborate on fine-tuning for similar downstream tasks. Different from the classic FL setting \cite{zhang2021parameterized, wang2023dafkd, lin2020ensemble}, we do not assume the data distribution on clients or the public data owned by the server to be the same. Our goal, in general, is to provide a framework for collaborative clients to fine-tune without access to full LLM or sharing local data directly. More importantly, the fine-tuned model can still achieve better performance than fine-tuning LLM locally with their local data exclusively. 

Although FedOT~\citep{kuang2023federatedscope} was developed for this objective, it could incur significant computational and communication costs, thereby suffering from the second challenge. In light of this challenge, we propose to integrate various parameter-efficient fine-tuning (PEFT) techniques into the proposed FL framework. Specifically, the server employs linear dropout to compress the LLM, integrates LoRA \cite{hu2021lora} to reduce the trainable parameters, and divides it into two components: an emulator and an adapter. The emulator retains a consistent representation of the raw model on the server's dataset, while the adapter assimilates domain-specific linguistic patterns from the clients' local datasets. Considering the significant distribution shift between the clients' datasets and the server's dataset, we separate the fine-tuning of these two components into two processes during FL training, i.e., the clients perform multiple local updates to fine-tune the adapter, and the server distill the emulator from the original LLM while aggregating the updated adapters from the clients. To this end, a bi-level optimization is formulated. 

This design, named \textbf{Fed}erated \textbf{Bi}-level \textbf{O}ffsite \textbf{T}uning (\algo), offers twofold advantages from the clients' perspectives. Firstly, instead of loading the complete model, clients load a compressed version with fewer layers, considerably reducing computation costs. Secondly, clients exclusively fine-tune the adapter, affecting only the last few layers of the LLM and thereby minimizing computation and communication expenses. 

\mypara{Contributions.} Throughout the paper, our contributions are highlighted as follows: 
\begin{itemize}[leftmargin=1em]
    \item We propose an algorithm \algo that avoids full model fine-tuning and significantly reduces the communication and computation overhead. To the best of our knowledge, this is the first work that addresses both the aforementioned two challenges in the federated LLM fine-tuning framework. With our proposed framework,  clients' data are ensured to be kept locally and computation and communication burden is significantly reduced. 
    \item We formulate a bi-level optimization problem that enables the LLM fine-tuning without access to the full model. By partitioning the compressed model into the adapter and the emulator, the emulator acts as a simulator of the original raw model, while the adapter adeptly learns domain-specific linguistic patterns with clients' local datasets. To this end, we realize that fine-tuning the compressed model is equivalent to the refinement of the counterpart of the complete LLM. 
    \item We conduct extensive experiments on LLaMA-2 for fine-tuning with three tasks, i.e., code generating, math problem solving, and question answering. The empirical studies also demonstrate that the proposed approach has significant improvement over all these tasks compared with the baseline approaches in terms of computation and communication overheads and final accuracy. 
\end{itemize}

\section{Preliminary} 

\subsection{Traditional FL Formulation} 

Consider there is an FL system with a total of $M$ clients, denoted by $[M]$. 
Each client $m \in [M]$ holds a local dataset $\mathcal{D}_m$. A client's local loss is defined as 
\begin{equation}
F_m(\param) := \frac{1}{|\mathcal{D}_m|} \sum_{(\dinput, \dtarget) \in \mathcal{D}_m} \loss\left(\model(\dinput; \param); \dtarget\right),    
\end{equation}
where $\model(\dinput; \param)$ is the output on a given model parameterized by $\param$ and an input $\dinput$. The loss function $\loss$ is defined on the model output and the ground truth $\dtarget$. In this dataset, we assume that the ground truth $\dtarget$ is part of the input $\dinput$, where a sequence of tokens in the input is used to predict the next token, and the ground truth is used to identify the part needing to be predicted by the model. Such a dataset is commonly adopted in previous works to fine-tune an LLM \cite{wei2021finetuned, ouyang2022training}. Then, based on the definition, a conventional FL system aims to find an optimal model across all clients, which is formulated as 
\begin{equation} \label{eq:fl_equation}
    \min_{\param \in \mathbb{R}^d} F(\param) = \sum_{m \in [M]} p_{m} F_m(\param),
\end{equation}
where $p_m = |\mathcal{D}_m| / |\mathcal{D}|$ for all $m \in [M]$, where $\mathcal{D}$ represents the entire training dataset, i.e., $\mathcal{D} = \cup_{m \in [M]} \mathcal{D}_m$. 
Generally, this problem can be optimized by different FL algorithms \cite{li2019convergence, karimireddy2020scaffold,wang2020tackling} repeating the following paradigm until convergence:
\begin{itemize}[leftmargin=1em]
    \item \textbf{Step 1:} At the beginning of each round $t$, the server broadcasts the trainable model parameters $\param^{(t)}$;
    \item \textbf{Step 2:} After receiving the model $\param^{(t)}$, each client $m \in [M]$ performs multi-step local updates on $\param^{(t)}$ to obtain $\param^{(t)}_m$;
    \item \textbf{Step 3:} The server collects the locally updated model parameters $\param^{(t)}_m$ from clients and aggregates them into a single global model $\param^{(t+1)}$ for next round. 
\end{itemize}

\mypara{Applying PEFT to federated LLM fine-tuning.}
The existing FL algorithms \cite{acar2020federated, wu2023anchor, he2023gluefl, wang2023fedcda, wang2022fedkc} are confronted with computation and communication bottlenecks when fine-tuning an LLM. To mitigate the limitations, researchers have extended existing parameter-efficient fine-tuning (PEFT) approaches to FL, named FedPEFT \cite{yi2023fedlora, zhang2023fedpetuning, sun2024improving}. These methods minimize the number of trainable parameters by introducing a PEFT module and keeping the original LLM parameters unchanged. By focusing local updates exclusively on the PEFT module rather than the entire model, these methods effectively reduce computational load and support larger batch sizes on a single GPU. Additionally, the FL server merely aggregates the updated parameters of a given model, thus obviating the need to transmit unchanged parameters and minimizing communication overheads. 

Nevertheless, FedPEFT is still confronted with the intrinsic challenge wherein clients face obstacles in loading an LLM due to its substantial computation prerequisites. For instance, the loading of a full-precision LLaMA-2-7B necessitates a memory capacity of no less than 28 GB. 

\subsection{Related Work}

The era of LLM poses the necessity of model privacy protection, where the details of LLM cannot be visible to the clients. To this end, \citet{xiao2023offsite} proposes a method named Offsite-tuning under the scenario where there is a server (a.k.a. LLM owner) and a client, while \citet{kuang2023federatedscope} extends this work to an FL version and names it as FedOT. They achieve model privacy protection by compressing the model, where only some layers are visible to the clients. However, these works require the preservation of a large number of layers to guarantee the performance, hindering the effectiveness of model privacy protection. In contrast, our work only discloses a few model parameters of the original LLM to the clients, i.e., the clients only know the adapter parameters that come from the original LLM, while the emulator parameters have been updated and different from the original LLM. Besides, neither offsite-tuning \cite{xiao2023offsite} nor FedOT \cite{kuang2023federatedscope} consider the difference between alignment data on the server and the fine-tuning data on clients. In contrast, the bi-level optimization problem proposed in our work naturally considers this factor and we design updating rules based on it.

Black-box is also a practical way to protect model privacy, where the clients access the LLM via an API, and they cannot fine-tune the LLM. Therefore, the optimization solely relies on prompt-based learning \cite{sordoni2023joint, li2021prefix, lester2021power, liu2023gpt}. In the context of FL, there are two typical works, namely, Fed-BBPT \cite{lin2023efficient} and FedBPT \cite{sun2023fedbpt}. These two works guarantee the model privacy in FL, but they should transmit the prompt together with the input to the LLM owner, leading to concerns about data privacy when the input contains sensitive information, violating the requirement of FL. In contrast, the proposed \algo will not lead to this concern because its training is fully on the clients such that the data are never shared with others.

\section{\algo} \label{sec: fedbiot}

Given that some clients may be unable to load a complete LLM, this section introduces an algorithm designed to enable these clients to fine-tune the LLM without requiring access to its full version. In other words, our goal is to refine the part of a compressed model that should yield performance comparable to fine-tuning its counterpart within a full model. To accomplish this, the server initially compresses the LLM and divides it into two distinct components, each serving specific functions. The first component, termed an emulator, is tasked with replicating the behavior of the uncompressed LLM. The second component, referred to as an adapter, focuses on adeptly acquiring domain-specific linguistic patterns from clients. Upon reintegrating the adapter into the uncompressed emulator, its performance should demonstrate significant improvement compared to the original LLM.

However, direct fine-tuning of the adapter on its models presents two significant limitations. Firstly, given that a single layer of a large language model (LLM) comprises millions of parameters, such as the decoder layer of LLaMA-2 with 202 million parameters, the adapter's parameter count is immense. This necessitates clients to possess powerful computational equipment to handle the fine-tuning of the layer. Additionally, transmitting the layer updates to the server poses another bottleneck, particularly in scenarios with unreliable network connections or limited bandwidth, hindering the smooth transmission of updates to the server.

To address these constraints, we integrate LoRA \cite{hu2021lora}, a PEFT module, into our proposed method. LoRA significantly reduces the number of tunable parameters, with a LoRA module for LLaMA-2 comprising 0.13 million trainable parameters, which is merely 0.06\% of the original layer's size. Consequently, the communication cost experiences a remarkable reduction of 99.94\% compared to transmitting a full layer.

\mypara{Organization.} In the subsequent sections, we will delve into the concrete details of the algorithm design. Specifically, Section \ref{subsec:preparation} illustrates how the compressed model is prepared. Following that, Section \ref{subsec:bilevel} discusses the problem formulation for the aforementioned objectives. On top of this, Section \ref{subsec:local_update} and Section \ref{subsec:aggregation} outline the detailed steps of the proposed algorithm, namely local updates and server aggregation, showcasing the seamless integration of LoRA modules. Full implementation of the pseudocode is given in Algorithm \ref{algo}. 

\subsection{Compressed Model Preparation: Linear Dropout} \label{subsec:preparation}

Suppose a pre-trained LLM has a total of $n$ layers of transformers. In the work, a repeatedly used operation is layer extraction, which extracts some layers out of the total $n$ layers of transformers to form a submodel. We denoted this by a function $\extract (\model, L)$, which means extracting the layers with indices in $L \subseteq [n]$ from the model $\model$. The function consists of the following three steps, and its pseudocode implementation of the first two steps is presented in Algorithm \ref{extract}. 

\mypara{Step 1: Identify the adapters in the original model.} 
We choose the bottom few layers\footnote{The bottom/last layers refer to the transformer decoders near the output, while the top/first layers refer to the part close to the input.} of the original LLM as the adapter. To be more specific, suppose the size of the adapter is $a$, and denote the adapter as $\adapter$. Therefore, $\adapter \leftarrow \extract(\model, L_{\adapter})$ with the layer indices $L_{\adapter} = \{n-a+1, \dots, n\}$. We denote $\param_{\adapter}$ as the parameters of $\adapter$. The $\param_{\adapter}$ are also the trainable parameters that can be fine-tuned by clients. The remaining part of the model is demoted as $\emulator^* = \model \setminus \adapter$. 

The choice of adapters brings two advantages. First, regarding the computation constraints of the clients, this proposed adapter is computation-efficient because it only needs to store the activations of transformers in the last few layers, leading to a lower memory cost. Second, as the adapter focuses more on domain-specific features, it is eco-friendly to spend the effort fine-tuning the last few layers. The conclusion is drawn from a well-known finding \citep{yosinski2014transferable} in neural networks that the first few layers tend to learn general features while the last layers encode specific ones.

\mypara{Step 2: Layer dropout to form emulator.} 
Inspired by the experimental results presented by \citet{xiao2023offsite}, we form an emulator by means of a uniform layer dropout \cite{sajjad2023effect} from the remaining part $\emulator^*$. Therefore, the emulator is a sub-model obtained as $\emulator \leftarrow \extract(\emulator^*, L_{\emulator})$. Denote there are $n_{\emulator^*}$ layers transformer in $\emulator^*$. The dropout rate of the emulator is denoted as $\beta = \frac{|L_{\emulator}|}{n_{\emulator^*}}$. For convenience, we call $\emulator$ as emulator and $\emulator^*$ as non-compressed emulator. Let $\param_{\emulator}$ and $\param_{\emulator^*}$ be the parameters of $\emulator$ and $\emulator^*$, respectively. 

After training, we can attain two combined models, namely, \underline{Adap}tor + \underline{Emu}lator (AdapEmu, i.e., $\emulator \circ \adapter$), and \underline{Adap}tor + \underline{Fu}ll (AdapFu, i.e., $\emulator^* \circ \adapter$). As \citet{xiao2023offsite} describes, AdapFu performs better than AdapEmu. These two models have different functionalities in real-world scenarios: AdapEmu is adopted if the input contains sensitive information that cannot be shared with the LLM owner, e.g., drafting a petition letter, while AdapFu is adopted when the users aim to have better generation results, e.g., solving a math problem. 

\begin{algorithm}[t]
\caption{\extract}\label{extract}
\begin{flushleft}
\textbf{Input:} pre-trained LLM $\model$ (layer index starts from 0), adapter size $s$, dropout rate $\beta$. 
\end{flushleft}
\begin{algorithmic}[1]
\State Get the size of model: $n \leftarrow |\model|$
\State Compute the number of layers in the compressed model, i.e., 
\begin{align*}
    n' \leftarrow \lfloor \beta (n-s) \rfloor
\end{align*}
\State Initialize non-compressed emulator $\emulator^* \leftarrow \{\model_{0}, \dots \model_{n-s-1}\}$, adapter $\adapter \leftarrow \{\model_{n-s}, \dots, \model_{n-1}\}$
\State Initialize emulator $\emulator \leftarrow \{\}$
\State Compute $stride \leftarrow (n-s-1) / (n'-1)$
\For{$j = 0, \dots, n'-1$}
    \State Append $\model_{\lfloor j \times stride \rfloor}$ to emulator, i.e., 
    \begin{align*}
        \emulator \leftarrow \emulator \cup \{\model_{\lfloor j \times stride \rfloor}\}
    \end{align*}
\EndFor
\State \Return $\param_{\adapter}, \param_{\emulator}, \param_{\emulator^*}$
\end{algorithmic}
\end{algorithm}

\begin{figure*}[t]
    \centering
    \includegraphics[width=\textwidth]{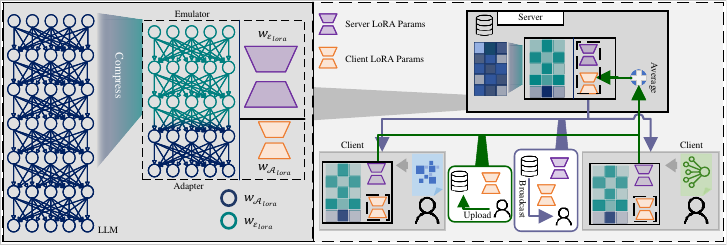}
    \caption{The Workflow of \algo during the FL training}
    \label{fig:workflow}
\end{figure*}

\newcommand{\lossfunc}{\mathcal{L}}
\mypara{Step 3: Pre-alignment.}
Before the FL training stage, we pre-align the emulator with the non-compressed one such that it can mimic the performance of the raw model. Assume there is a public dataset $\mathcal{D}_{public}$ available on the server, which consists of a bunch of data $(\dinput, \dtarget)$, representing the input and the ground truth, respectively. Therefore, in the rest of the section, we assume the input $\dinput$ contains an attention mask that can identify the ground truth. 

Instead of training the compressed model with the ground truth, we utilize knowledge distillation \cite{hinton2015distilling} to transfer the general linguistic patterns from the original LLM to the compressed one by tuning the emulator $\emulator$. In specific, we ensure the emulator generates representations that have subtle differences from the non-compressed emulator with the given input on the ground truth part. To this end, we aim to minimize the following $\ell_2$-norm: 
\begin{align}
    \lossfunc_{repr} &= \left\| \emulator\left(\dinput ; \param_{\emulator}\right) - \emulator^*\left(\dinput ; \param_{\emulator^*}\right)  \right\|_2^2
\end{align}
Additionally, we ensure the compressed model has consistent final outputs of the original LLM on the ground truth by minimizing the following KL divergence: 
\begin{align}
    \lossfunc_{kd} &=  D_{KL} \left( \model(\dinput ; \{\param_{\adapter}, \param_{\emulator^*}  \}) \| \model(\dinput ; \{ \param_{\adapter}, \param_{\emulator}\}) \right)
\end{align}
In a nutshell, we optimize the emulator $\emulator$ by finding the optimal parameters for the following equation on the public dataset 
\begin{align} \label{eq:opt_emu}
    \min_{\param_{\emulator}} \frac{1}{|\mathcal{D}_{public}|} \sum_{\substack{\dinput \in \mathcal{D}_{public}}} \lossfunc_{repr} + \lambda \lossfunc_{kd}
\end{align}

Let the optimal emulator $\emulator$ for Equation \eqref{eq:opt_emu} be $\emulator_{init}$ with the parameter of $\param_{\emulator_{init}}$. Denote the selected adapter $\adapter$ with the parameter of $\param_{\adapter_{init}}$. To this end, we distribute a compressed model to the clients with the initial parameters of $\{\param_{\adapter_{init}}, \param_{\emulator_{init}}\}$. To reduce the computation and communication costs, we incorporate LoRA \cite{hu2021lora} for the adapter $\adapter$ and the emulator $\emulator$, denoted as $\adapter_{lora}$ and $\emulator_{lora}$, respectively. 

Before diving into the details of the proposed \algo, we briefly go through the workflow as described in Figure \ref{fig:workflow}. The figure visually presents the workflow of the federated learning process of our proposed \algo, including the local updates on clients (Section \ref{subsec:local_update}) and the aggregation on the server (Section \ref{subsec:aggregation}). At the beginning of clients' fine-tuning, the server broadcasts the adapter $\adapter_{lora}$ and the emulator $\emulator_{lora}$ to the clients. Subsequently, the clients perform multiple local updates to fine-tune the adapter $\adapter_{lora}$ with their local datasets. After the local updates, the client uploads the adapter $\adapter_{lora}$ to the server, and the server thereby aggregates the adapters. To ensure that the emulator is still able to reproduce the behavior of the uncompressed LLM, the server fine-tunes the emulator $\emulator_{lora}$ with the public dataset. Finally, the server distributes the updated parameters to the clients and launches a new round of training. 

\subsection{Formulation of Bi-level Optimization} \label{subsec:bilevel}

As discussed in Section \ref{subsec:preparation}, we compress and divide the LLM into two parts, namely, an adapter and an emulator. These two components are designated to satisfy the following objectives: 
\begin{itemize}[leftmargin=1em]
    \item \textbf{Emulator} should be tuned towards perfectly imitating the non-compressed part in the full model, especially in extracting and encoding information on the server's datasets.
    \item \textbf{Adapter} should be able to digest the output of the emulator efficiently and should be encoded with the knowledge from clients' datasets effectively. 
\end{itemize}

Define $\param_{\adapter} = \{\param_{\adapter_{init}}, \param_{\adapter_{lora}}\}$, $\param_{\emulator} = \{\param_{\emulator_{init}}, \param_{\emulator_{lora}}\}$ to integrate the LoRA parameters while the initial parameters for the adapter and the emulator remain unchanged during the training. Toward the goal, we formulate the objectives as a bi-level optimization problem: 
\begin{align}
    &\min_{\param_{\adapter_{lora}}} \sum_{m \in [M]} p_m F_m\left(\{ \param_{\adapter}, \param_{\emulator}\}\right) + \frac{\epsilon}{2} \left\|\hat{\param}_{\adapter_{lora}} - \hat{\param}_{\adapter_{lora}}^{(t)}\right\|_2^2 \label{eq:client_training} \\ 
    s.t. \quad &\param_{\emulator_{lora}} \in \arg\min_{\param_{\emulator_{lora}}} \frac{1}{|\mathcal{D}_{public}|} \sum_{\substack{\dinput \in \mathcal{D}_{public}}} \mathcal{L}(\dinput), \nonumber \\
    & \mathcal{L}(\dinput) \overset{\triangle}{=}  \left\| \emulator\left(\dinput ; \param_{\emulator}\right) - \emulator^*\left(\dinput ; \param_{\emulator^*}\right)  \right\|_2^2 \nonumber \\
    & \qquad\quad + \lambda \cdot D_{KL} \left( \model(\dinput ; \{\param_{\adapter}, \param_{\emulator^*}  \}) \| \model(\dinput ; \{ \param_{\adapter}, \param_{\emulator}\}) \right) \label{eq:emu_align}
\end{align}
where $\param_{\adapter_{lora}}^{(t)}$ is the adapter LoRA received at the beginning of each communication round, $\hat{\param}_{\adapter_{lora}}$ reconstructs for the same size of the adapter $\param_{\adapter}$. $\mathcal{D}_{public}$ represents the public dataset on the server, which can be unlabeled. $D_{KL}(\cdot \| \cdot)$ is the KL divergence between two logits. $\epsilon$ and $\lambda$ are hyperparameters.

\mypara{The upper-level objective (Equation~\eqref{eq:client_training}).}
The upper-level objective function consists of two terms. The first term represents the loss of the model on local clients' data, with the current emulator and adapter. It follows a classic weighted average loss in FL to balance the loss of different clients' heterogeneous local data. The goal of introducing this term is straightforward: by minimizing the loss of the first term, we expect the emulator-adapter combination to be improved on the local training set. The second term is a regularization of the adapter component to ensure it will be within a reasonable distance from the synchronized and broadcast adapter at the beginning of each communication round. Enforcing a restriction on the adapter's change can reduce the difference of losses for the emulator distillation after locally adapter are tuned locally on clients, so it can help the convergence of emulator distillation.

\mypara{The lower-level objective (Equation~\eqref{eq:emu_align}).} The first term in the constraint is the $\ell_2$-norm difference between the activation output by the emulator and the full model. The second term is the KL divergence between the output of output distribution of the full model-adapter combination and the emulator-adapter. Although only the emulator is trainable to minimize the loss of these two terms, these two terms provide different optimization meaning for the emulator. The first term encourages the emulator to provide activations as close as possible to the full model, \emph{excluding} the effect of the adapter. The second term ensures the emulator can provide output distributions close to the one when the full model with adapters is added on.

\mypara{Discussion.} The introduced algorithm can optimize the bi-level problems (i.e., Equation \eqref{eq:client_training} and \eqref{eq:emu_align}) to an equilibrium point for both adapter and emulator. This is because when we optimize the adapter, the fixed emulator constrains its updates, and vice versa, and thereby, the emulator and adapter are distilled or trained interchangeably. At this equilibrium, the emulator can more faithfully extract and encode the information for the clients' dataset and benefit from the training of the adapter in reverse.

Additionally, \algo does not require the design of an emulator to follow linear dropout. Instead, this is a general framework that compresses an LLM and divides it into two components: an emulator and an adapter. There are numerous designs for the emulator, but they share the same objective where the emulator simulates the non-compressed part of an LLM. For simplicity, we follow offsite-tuning \cite{xiao2023offsite} and prepare the emulator by means of uniform layer dropout \cite{sajjad2023effect} to demonstrate the effectiveness of \algo.

\begin{algorithm}[t]
\caption{\algo}\label{algo}
\begin{flushleft}
\textbf{Input:} learning rate $\eta$, local updates $K$, global model alignment steps $E$, strength of full model alignment $\lambda$, local update regularization $\epsilon$, total communication rounds $R$, pre-trained LLM $\model$ with parameter $\param$, adapter size $s$, dropout rate $\beta$, number of clients $M$.
\end{flushleft}
\begin{algorithmic}[1]
\State $\param_{\adapter}, \param_{\emulator}, \param_{\emulator^*} \leftarrow \extract(\model, s, \beta)$ \Comment{See Algo. \ref{extract} for details}
\For{$t = 0, \dots, R-1$}
    \For{$e = 0, \dots, E-1$}
        \State Randomly sample $(\dinput, \dtarget)$ from the public dataset $\mathcal{D}_{public}$
        \State Optimize $\param_{\emulator_{lora}}$ with respect to Equation \eqref{eq:emu_align}
    \EndFor
    \State Communicate $\{\param_{\adapter_{lora}}, \param_{\emulator_{lora}}\}$ with clients $m \in [M]$
    \For{$m \in [M]$ \textbf{in parallel}}
        \State Initialize $\param_{\adapter_{lora}}^{(t)}$, $\param_{\adapter_{lora}, m}$, and $\param_{\emulator}$ using Equation \eqref{eq:init}
        \For{$k = 0, \dots, K-1$}
            \State Compute a gradient $g$ using Equation \eqref{eq:comp_grad}
            \State Update local model $\param_{\adapter_{lora}, m}$ using Equation \eqref{eq:local_optim}
        \EndFor
        \State Communicate $\param_{\adapter_{lora}, m}$ with the server 
    \EndFor
    \State Update $\param_{\adapter_{lora}}$ using Equation \eqref{eq:global_aggregation}
\EndFor
\State \Return AdapEmu $\{\param_{\adapter}, \param_{\emulator}\}$, AdapFu $\{\param_{\adapter}, \param_{\emulator^*}\}$
\end{algorithmic}
\end{algorithm}

\subsection{Client Updates} \label{subsec:local_update}

During the local updates, the clients barely fine-tune the parameters of the adapter $\adapter$ while fixing the parameters of the emulator $\emulator$. By enabling LoRA, the LoRA of the adapter will get updated, and therefore, the clients should upload the updated $\param_{\adapter_{lora}}$ to the server after the local fine-tuning ends. 

Consider client $i \in [M]$ performs the local updates at $t$-th round. Before optimizing the adapter locally, the client receives the updated emulator $\param_{\emulator_{lora}}$ and adapter $\param_{\adapter_{lora}}$ from the client, and we denote them by 
\begin{align} \label{eq:init}
    \param_{\adapter_{lora}}^{(t)} \leftarrow \param_{\adapter_{lora}}, \quad \param_{\adapter_{lora}, m} \leftarrow \param_{\adapter_{lora}}, \quad \param_{\emulator} \leftarrow  \{\param_{\emulator_{init}}, \param_{\emulator_{lora}}\}
\end{align}
Suppose the client performs the local update for $K$ times. In each local update, we solely optimize $\param_{\adapter_{lora}, m}$, a LoRA module of the adapter. Therefore, based on Equation \eqref{eq:client_training}, the gradient w.r.t. $\param_{\adapter_{lora}, m}$ should be
\begin{align} \label{eq:comp_grad}
    g \leftarrow \nabla_{\param_{\adapter_{lora}, m}} F_m\left(\{ \param_{\adapter}, \param_{\emulator}\}\right) + \epsilon \left(\hat{\param}_{\adapter_{lora}, m} - \hat{\param}_{\adapter_{lora}}^{(t)}\right)^{T} \left(\frac{\partial \hat{\param}_{\adapter_{lora}, m}}{\partial \param_{\adapter_{lora}, m}}\right)
\end{align}
where $\param_{\adapter} = \{\param_{\adapter_{init}}, \param_{\adapter_{lora}, m}\}$ in the above formula. Let $\textsc{Optim}()$ be the optimizer (e.g., SGD and AdamW \cite{loshchilov2017decoupled}) that updates the model parameters, and $\eta$ be the learning rate. Therefore, in each local update, the local model is updated for 
\begin{align} \label{eq:local_optim}
    \param_{\adapter_{lora}, m} \leftarrow \textsc{Optim}(\param_{\adapter_{lora}, m}, g, \eta)
\end{align}
After finishing the local update, the client $i$ sends $\param_{\adapter_{lora}, m}$ to the server. 

\subsection{Model Aggregation} \label{subsec:aggregation}

During the server aggregation, the server performs the weighted average to update the adapters $\adapter$ and fine-tune the emulator $\emulator$. By enabling the LoRA, only the parameters $\param_{\adapter_{lora}}$ in the adapter and $\param_{\emulator_{lora}}$ in the emulator are updated, while the rest (i.e., $\param_{\adapter_{init}}$ and $\param_{\emulator_{init}}$) remain unchanged. 

First, the server collects a set of updated LoRAs of the adapter, i.e., $\left\{\param_{\adapter_{lora}, m}\right\}_{m \in [M]}$ from the clients. Based on the definition of Equation \eqref{eq:client_training}, the server performs weighted aggregation via
\begin{align} \label{eq:global_aggregation}
\param_{\adapter_{lora}} \leftarrow \sum_{m \in [M]} p_m \param_{\adapter_{lora}, m}
\end{align}

After the weighted averaging, the server distills the emulator $\emulator$ from the non-compressed $\emulator^*$ and the updated adapter $\adapter$ using the public dataset. Therefore, we fine-tune the emulator following Equation \eqref{eq:emu_align}, which updates the LoRA of the emulator $\param_{\emulator_{lora}}$. 

\section{Experiments} \label{sec:exp}

\subsection{Experimental Setup} 

This section discusses the implementation of our experiments, covering details such as the model utilized and evaluation metrics. The code is now available at \url{https://github.com/HarliWu/FedBiOT}. 

\mypara{Model and computation environment.} 
The experiments utilize LLaMA-2-7B, an open-source pre-trained LLM maintained by Meta and released in July 2023 \cite{touvron2023llama}. Preceding this, the model's first generation was introduced in February 2023 \cite{touvron2302llama}. This model supports a maximum of 4096 input tokens and consists of 32 hidden layers with a total of 6.7 billion parameters. The experimental setup involves machines equipped with Nvidia A100 GPU cards, Intel Xeon Platinum 8369B CPUs, and a 512GB RAM configuration. 

\begin{table*}[t]
\renewcommand{\arraystretch}{1.1}
    \centering
    \caption{Dataset details for LLM training and evaluation}
    \vspace{-5px}
    \begin{tabular}{lccccccccc}
        \Xhline{1.5pt}
         Task & \makecell{Training\\Dataset} & \makecell{\# training\\samples} & \# clients & Partition Rules & Max. & Min. & Std. & \makecell{Test\\Dataset} & \makecell{\# test\\samples} \\\hline
         Math Problem Solving & GSM-8K & 7473 & 3 & i.i.d. & 2491 & 2491 & 0 & GSM-8K & 1319 \\
         Code Generation & Rosetta & 7954 & 9 & Prog. Lang. & 1172 & 439 & 236.94 & HumanEvalX & 656 \\
         Question Answering & Dolly & 15015 & 8 & Category & 3611 & 711 & 795.06 & Helm & NA \\
         Public Dataset & Alpaca & 52002 & \multicolumn{7}{c}{------------------------------------------------------------------------------------------}\\
         \Xhline{1.5pt}
    \end{tabular}
    \label{tab:datasets}
\end{table*}

\mypara{Datasets and Tasks.} In the experiments, we use the benchmark datasets and tasks in \citep{kuang2023federatedscope} to train and evaluate the LLM on three different NLP tasks, covering math problem-solving, code generation, and question-answering: 
\begin{itemize}[leftmargin=1em]
    \item For \textbf{math problem-solving}, we split the GSM-8K training dataset \cite{cobbe2021training} ensuring i.i.d. across three clients, and we assess the updated model using the GSM-8K test dataset. 
    \item For \textbf{code generation}, we fine-tune the model with the Rosetta dataset \cite{codealpaca}, which is partitioned across the programming languages, and a total of nine clients separately hold the data from nine different programming languages. Regarding its evaluation, we utilize HumanEvalX \cite{zheng2023codegeex}, an extension of a coding evaluation dataset \cite{chen2021evaluating} that requires the model to fill in the code for a given problem in the required programming language (i.e., C++, GO, Java, Python). 
    \item For \textbf{question answering}, the model is trained on dolly-15K \cite{DatabricksBlog2023DollyV2}, which is partitioned into 8 clients based on the categories of the questions, and we evaluate the new model with the selected tasks on HELM \cite{liang2022holistic}. 
\end{itemize}
Table \ref{tab:datasets} gives a detailed description of these three tasks. As Section \ref{sec: fedbiot} mentions, the server will perform the emulator alignment during the model aggregation. Then, we use the Alpaca dataset \cite{alpaca} as the public dataset for the server to do the emulator alignment for all three NLP tasks.

\mypara{Implementation.} This work is built upon an open-source federated learning platform named FederatedScope \cite{xie2022federatedscope}. The training data are reformatted following the predesigned instructions \cite{codealpaca, zhang2023towards}. 

Different from \citep{xiao2023offsite, kuang2023federatedscope}, we regard the last two and the last four decoders as the adapter. The experiments consider two dropout rates, i.e., $\beta \in \{0.2, 0.5\}$, and we obtain the emulators with layer dropout following~\citet{xiao2023offsite}. Without special annotation, we use the following local training setting: in each communication round, each client performs 30 local updates, and the batch size of every local update is 10. Before launching the FL training, we fine-tune the emulator for 500 iterations to generate a distilled emulator $\emulator$ towards minimizing the loss of Equation \eqref{eq:emu_align}. During the FL training, the server takes 10 iterations to align the emulator $\emulator$ with $\emulator^*$ between two successive communication rounds after aggregating local adapters with FedAvg~\cite{li2019convergence}. These experiments run for 500 communication rounds, and we report the results based on the fine-tuned LLM obtained at the 500th round. During the training, we only fine-tune the adapter in the clients' local update procedures, and we update the emulator on the server side. In other words, other parts of the pre-trained model, such as word embeddings, are frozen during the training. 

\mypara{LoRA, Optimizers and Hyperparameters.} We add the LoRA to all decoder layers in the adapter and the emulator by setting the rank to 8 and the alpha to 16. We use AdamW as an optimizer to solve Equation \eqref{eq:client_training} and \eqref{eq:emu_align} on the clients (for the adapters) and the server (for the emulators), respectively. We search for the best learning rate in $\{1\times10^{-5}, 3\times10^{-5}, 5\times10^{-5}, 8\times10^{-5}, 1\times10^{-4}\}$. We set the momentum for $(0.9, 0.95)$. As for other hyperparameters related to the optimizer, we use the default setting. Furthermore, we also conduct grid search for \algo-specific hyperparameters, i.e., $\epsilon$ and $\lambda$. Throughout the experiments, we demonstrate the result of the best hyperparameter combination. To avoid randomness, we utilize three different random seeds and report the averaged results. 

\mypara{Baselines.} Offsite-tuning is the only method that satisfies the constraints that fine-tuning without access to full model. \citet{xiao2023offsite} introduces a single-client offsite-tuning, while \citet{kuang2023federatedscope} extends it to an FL version (i.e., FedOT). We apply offsite-tuning with one single client, where all data are loaded to the client. As FedOT supports FL, we reproduce the algorithm to work on the FL tasks. In terms of the setting of the adapters and the emulators, both Offsite-tuning and FedOT treat the first two and the last two decoders as the adapter. To enable the parameter-efficient fine-tuning for both baselines, we add LoRA to both baselines, the same as the setting adopted by \algo. 

\mypara{Evaluation Metric.} In the experiments, we report the results on two models, i.e., AdapEmu and AdapFu, as defined in Section \ref{subsec:preparation}. The evaluation metrics for each task follow ~\citet{kuang2023federatedscope}, and the detailed description is given in Appendix \ref{apdx:eval}.

\begin{table}
\renewcommand{\arraystretch}{1.2}
    \centering
    \caption{Test accuracy on math problem-solving task under different dropout rates}
    \vspace{-5px}
    \resizebox{\linewidth}{!}{
    \begin{tabular}{llcc}
        \Xhline{1.5pt}
        \makecell[l]{Dropout\\Rate ($\beta$)} & Methods & AdapEmu & AdapFu 
        \\\hline
        $\beta=0.0$ & Few-shot CoT & NA & 13.42\% (177/1319) \\\hdashline
        
        \multirow{4}{*}{$\beta=0.2$} & Offsite-tuning & 3.03\% (40/1319) &  9.93\% (131/1319) \\
        & FedOT & 2.43\% (32/1319) & 10.16\% (134/1319) \\
        & \algo (Adapter 2) & \textbf{3.71\% (49/1319)} & 15.16\% (200/1319) \\
        & \algo (Adapter 4) & 3.41\% (45/1319) & \textbf{15.23\% (201/1319)} \\ \hline
        
        \multirow{4}{*}{$\beta=0.5$} & Offsite-tuning & \textbf{2.27\% (30/1319)} &  7.58\% (100/1319) \\
        & FedOT & 1.90\% (25/1319) & 7.51\% (99/1319) \\
        & \algo (Adapter 2) & 2.05\% (27/1319) & 11.83\% (156/1319) \\
        & \algo (Adapter 4) & 1.82\% (24/1319) & \textbf{14.03\% (185/1319)} \\ 
        \Xhline{1.5pt}
    \end{tabular}
    }
    \label{tab:gsm8k}
    \vspace{-10px}
\end{table}

\begin{table*}[t]
\renewcommand{\arraystretch}{1.1}
    \centering
    \caption{Pass@1 (\%) and Pass@10 (\%) in code generation task at various rounds when dropout rate is 0.2}
    \vspace{-5px}
    \begin{tabular}{lcccccccccccc}
        \Xhline{1.5pt}
          \multirow{2}{*}{Method} & \multirow{2}{*}{Model} & \multicolumn{2}{c}{C++} && \multicolumn{2}{c}{GO} && \multicolumn{2}{c}{Java} && \multicolumn{2}{c}{Python} \\ \cline{3-4} \cline{6-7} \cline{9-10} \cline{12-13}
          &  & Pass@1 & Pass@10 && Pass@1 & Pass@10 && Pass@1 & Pass@10 && Pass@1 & Pass@10 \\\hline
         
         \multirow{2}{*}{Offsite-tuning} & AdapEmu & 3.99 & \textit{6.45} && 1.80 & 2.44 && 5.64 & 6.09 && 5.01 & 6.38 \\
         & AdapFu & 8.78 & 10.82 && 4.94 & 6.63 && 9.57 & 12.81 && 13.19 & 17.32 \\\hline 
         
         \multirow{2}{*}{FedOT} & AdapEmu & 2.50 & 4.89 && 1.86 & 3.05 && 5.00 & 5.49 && 4.91 & 6.83 \\
         & AdapFu & 8.60 & 11.36 && 5.95 & 7.11 && 6.30 & 9.42 && 12.23 & 13.58 \\\hline 

         \multirow{2}{*}{\makecell[l]{\algo\\(Adapter 2)}} & AdapEmu & \textit{4.82} & 6.43 && \textit{3.57} & \textit{4.85} && \textit{5.92} & \textit{6.36} && 4.97 & \textit{6.95} \\
         & AdapFu & \textbf{9.76} & \textbf{14.18} && \textbf{9.97} & \textbf{13.29} && \textbf{12.93} & \textbf{16.28} && \textbf{14.91} & \textbf{19.77} \\\hline 

         \multirow{2}{*}{\makecell[l]{\algo\\(Adapter 4)}} & AdapEmu & 3.20 & 4.57 && 2.20 & 2.44 && 4.91 & 5.73 && \textit{5.43} & 6.10 \\
         & AdapFu & 9.12 & 13.41 && 8.02 & 11.08 && 11.28 & 13.10 && 14.57 & 18.41 \\\Xhline{1.5pt} 
    \end{tabular}
    \label{tab:code_2}
\end{table*}

\begin{table*}[t]
\renewcommand{\arraystretch}{1.1}
    \centering
    \caption{Pass@1 (\%) and Pass@10 (\%) in code generation task at various rounds when dropout rate is 0.5. We do not show AdapEmu's performance because it struggles to generate meaningful codes, accounting for its small size.}
    \vspace{-5px}
    \begin{tabular}{lcccccccccccc}
        \Xhline{1.5pt}
         \multirow{2}{*}{Method} & \multirow{2}{*}{Model} & \multicolumn{2}{c}{C++} && \multicolumn{2}{c}{GO} && \multicolumn{2}{c}{Java} && \multicolumn{2}{c}{Python} \\ \cline{3-4} \cline{6-7} \cline{9-10} \cline{12-13}
          &  & Pass@1 & Pass@10 && Pass@1 & Pass@10 && Pass@1 & Pass@10 && Pass@1 & Pass@10 \\\hline
         
         Offsite-tuning & AdapFu & 5.30 & 7.26 && 3.32 & 7.55 && 4.61 & 5.33 && 8.75 & 10.26 \\\hline 
         
         FedOT & AdapFu & 4.92 & 7.33 && 5.00 & 8.33 && 3.86 & 4.37 && 7.33 & 8.91 \\\hline 

         \algo (Adapter 2) & AdapFu & \textbf{7.71} & \textbf{11.84} && \textbf{7.68} & \textbf{10.01} && \textbf{9.51} & \textbf{14.34} && 13.29 & \textbf{16.87} \\\hline 

         \algo (Adapter 4) & AdapFu & 5.03 & 11.09 && 6.25 & 8.47 && 7.41 & 13.32 && \textbf{13.54} & 16.74 \\\Xhline{1.5pt} 
    \end{tabular}
    \label{tab:code_5}
\end{table*}

\subsection{Quantitative Evaluation on i.i.d. Data}

We demonstrate the experimental results of GSM-8K provided in Table \ref{tab:gsm8k} and highlight the worth-noted phenomenon when the data are i.i.d. across the clients.  

A notable phenomenon observed in the table is that AdapEmu significantly falls behind AdapFu, particularly at a low dropout rate (i.e., $\beta=0.2$). To explain this, we examine the accuracy of the LLaMA-2 model with a dropout rate of 0.2, which is 2.12\% without fine-tuning and increases to 2.43\% after fine-tuning the emulator with a public dataset. The performance gap between AdapEmu and AdapFu can be attributed to layer dropout, which reduces the size of the LLM and subsequently impacts its performance. Additionally, this result highlights the difficulty of accurately reproducing the non-compressed parts with the emulator. Fortunately, all methods improve AdapEmu's performance compared to the version without fine-tuning.

When we take a look at the proposed \algo at different adapters' sizes, we notice that \algo with adapter 4 achieves better performance than that with adapter 2 under the AdapFu setting. As we know, a larger adapter has more trainable parameters, and therefore, it can easily absorb the knowledge from the downstream tasks. Note that the performances of these two adapter settings have subtle differences under AdapEmu, meaning that their emulator achieves very similar effects to the non-compressed emulator. When we plug the adaptor back into the non-compressed emulator, the adapter with more trainable parameters obviously can achieve a better performance. 

When comparing our proposed model with the baselines, we can notice a significant dominance in performance, especially in the AdapFu setting. More specifically, when the dropout rate becomes larger, the performance of AdapFu with \algo decreases more mildly in contrast to other baselines. This is thanks to two factors: 1) the regularization term ensures the adapters will not change dramatically; 2) the on-the-fly distillation of the emulator with mixed losses can work better with clients' data. Although the other two baselines use a public dataset to achieve similar functionality, the deterioration may still occur due to the data domain shift and the significant information loss.

\subsection{Quantitative Evaluation on non-i.i.d. Data} 

According to Table \ref{tab:datasets}, code generation and question answering are two tasks split in non-i.i.d. styles. In this section, we evaluate our proposed \algo when it trains an LLM with a non-i.i.d. dataset. It is worth noting that the evaluation task could be either in-distribution or out-of-distribution to the training dataset. 

\mypara{Code generation.} Table \ref{tab:code_2} and \ref{tab:code_5} illustrate the best results in different programming languages based on different hyperparameter settings. Let us take a look at the results of the \algo at different adapter sizes. Apparently, \algo with two layers of adapter constantly outperforms \algo with four under both AdapEmu and AdapFu. This conclusion is different from the one when an LLM is trained with an i.i.d. dataset. The discrepancy can be attributed to the clients' objectives: under i.i.d. datasets, a larger adapter size benefits training by absorbing downstream linguistic patterns uniformly. Conversely, with non-i.i.d. datasets, clients are biased towards their local optima, where the emulator's effect becomes crucial. 

When comparing our proposed algorithm with the baselines, we notice a distinct dominance in AdapFu across all programming languages. In particular, when the dropout rate is 0.5, we can achieve up to 6\% improvement over other baselines in terms of Pass@1, and up to 10\% improvement of Pass@10. Notably, the most distinct dominance can be witnessed under the ``column'' of Java in Table \ref{tab:code_5}. 

\begin{figure*}[t]
\centering
\subfloat{
\centering
\includegraphics[width=.8\textwidth]{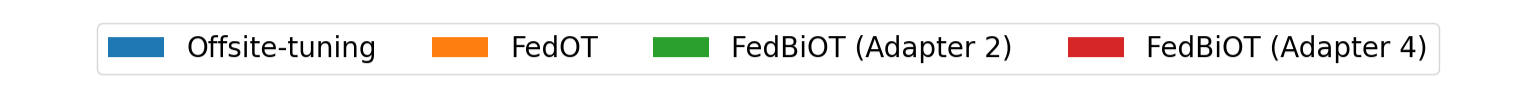}
}\\\vspace{-10px}
\setcounter{subfigure}{0}
\subfloat[AdapEmu (Dropout rate 0.2)]{\label{subfig:adapemu_dp2}
\centering
\includegraphics[width=0.33\textwidth]{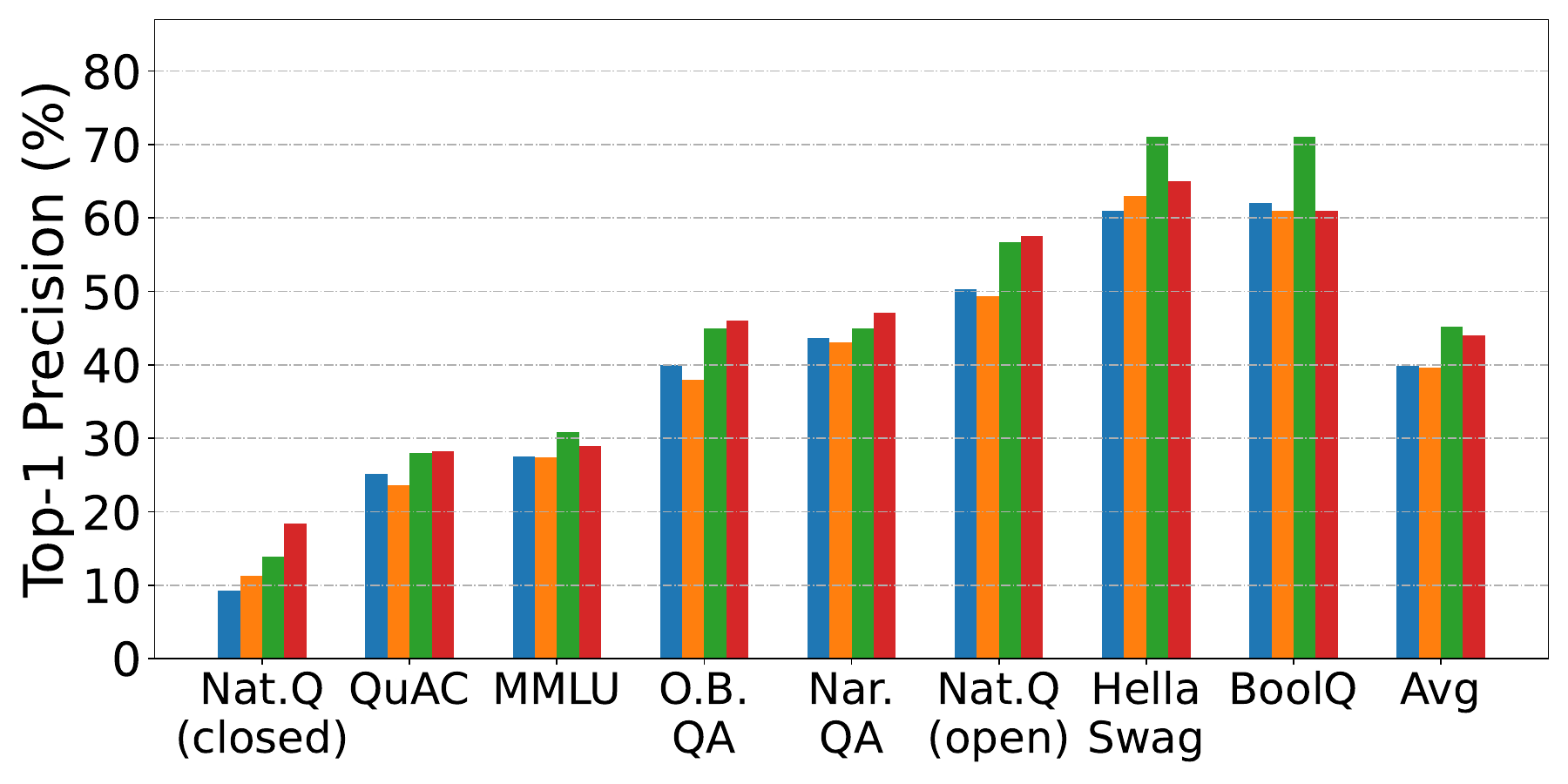}
}%
\subfloat[AdapFu (Dropout rate 0.2)]{\label{subfig:adapfu_dp2}
\centering
\includegraphics[width=0.33\textwidth]{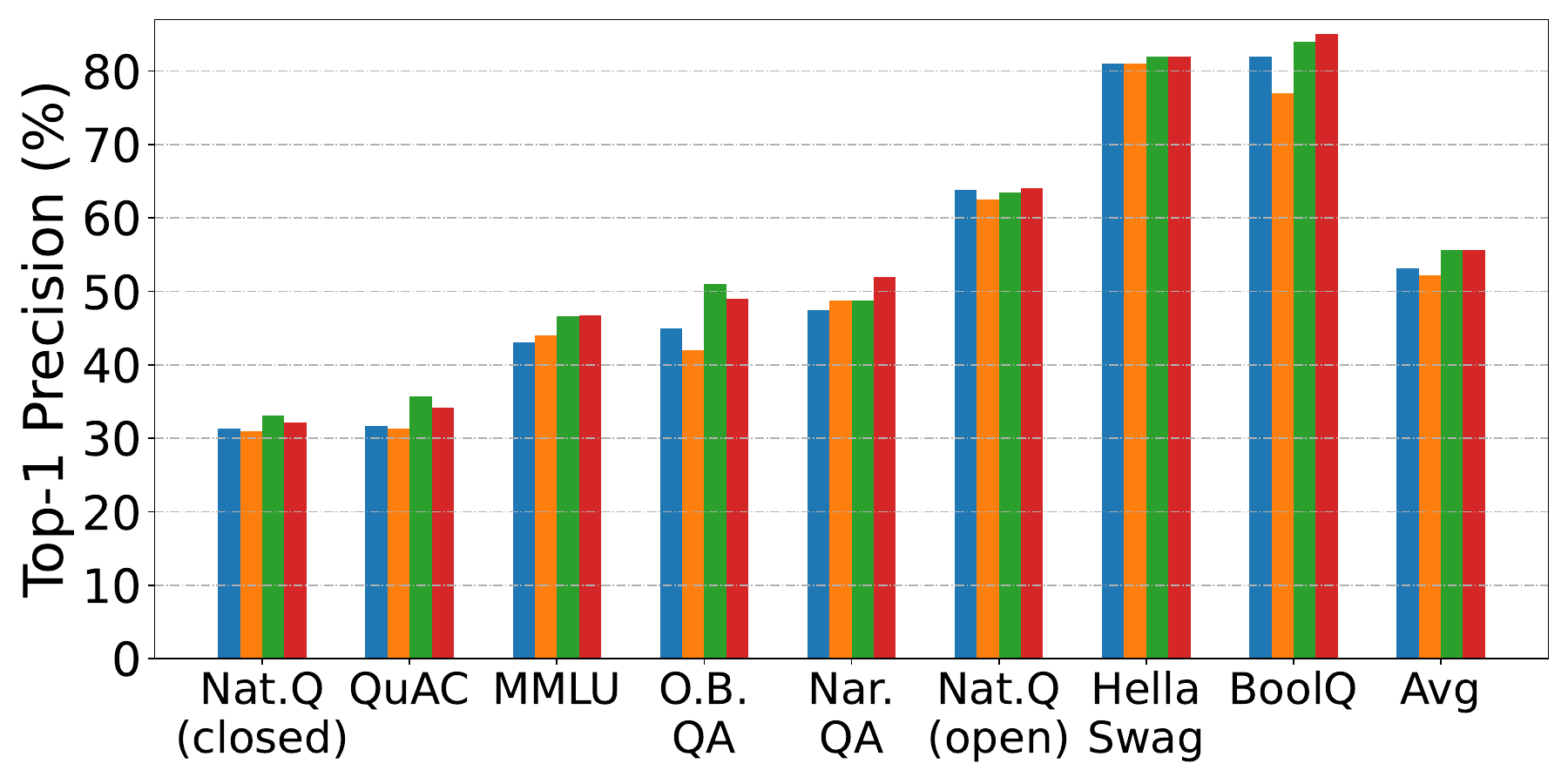}
}%
\subfloat[AdapFu (Dropout rate 0.5)]{\label{subfig:adapfu_dp5}
\centering
\includegraphics[width=0.33\textwidth]{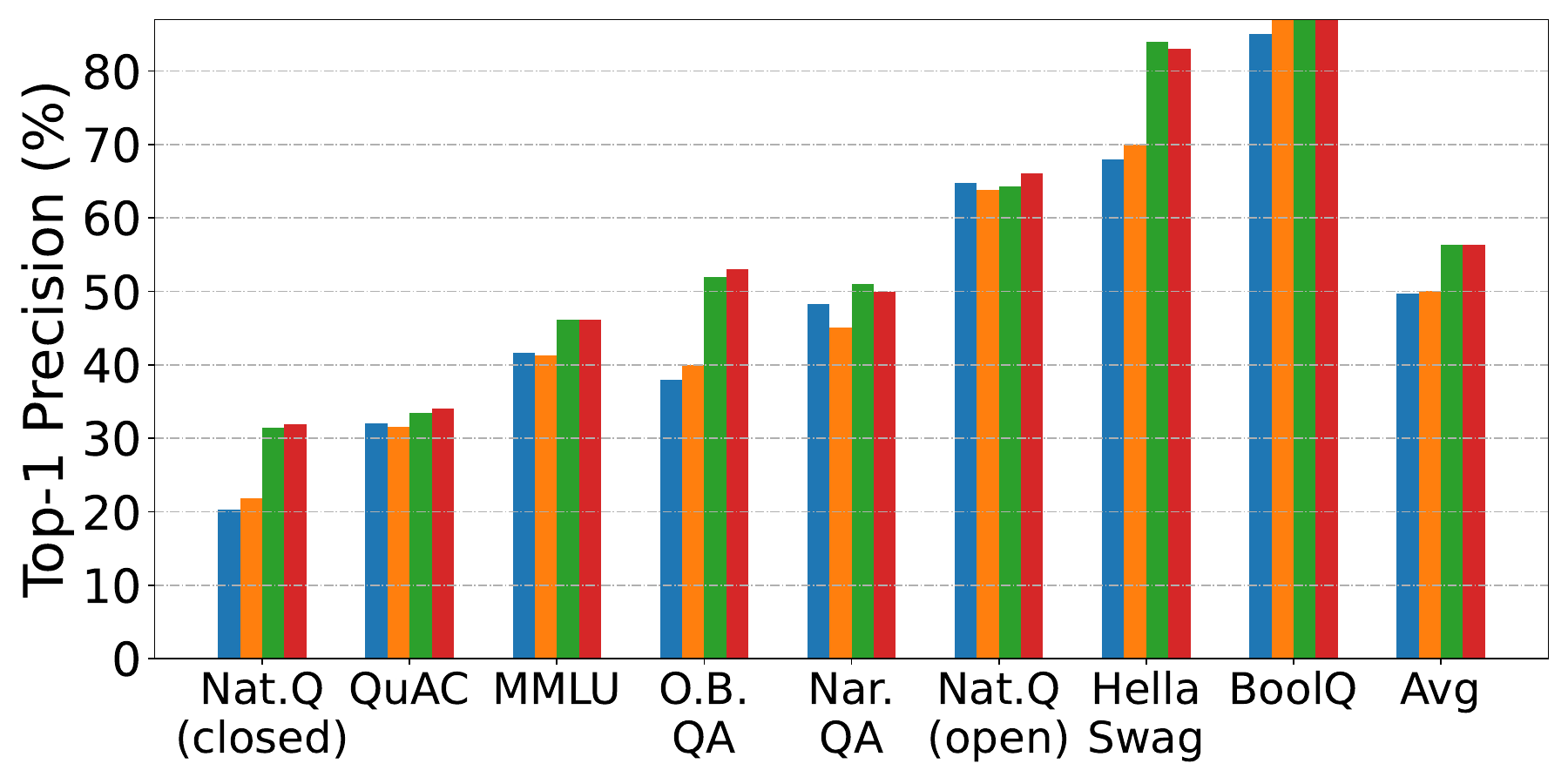}
}%
\centering
\vspace{-5px}
\caption{Test accuracy in eight types of question-answering tasks (Left to right: Natural Questions (closed-book), QuAC, MMLU, OpenbookQA, NarrativeQA, Natural Questions (open-book), HellaSwag, BoolQ) and the average accuracy under different baselines (bars from left to right: Offsite-tuning, FedOT, \algo (Adapter 2), \algo (Adapter 4)) and different dropout rates. }
\label{fig:helm_2}
\end{figure*}

\mypara{Question Answering.} Figure \ref{fig:helm_2} shows the evaluation results using the HELM benchmark while we train the LLM with Dolly-15K. Generally speaking, \algo (Adapter 2) performs significantly better than Adapter 4 in some tasks in terms of AdapEmu. As both AdapEmu have the same number of layers, this result exhibits the importance of the emulator, i.e., the model with a larger emulator can achieve leading performance. To some extent, this result supports our previous conclusion that an emulator plays a more important role than an adapter in a non-i.i.d. task. As for AdapFu, the performance difference is trivial between the two adapter sizes. 

The proposed algorithm outperforms offsite-tuning and FedOT in most datasets, which is consistent with the findings in other training tasks. The dominance of AdapFu becomes more pronounced as the dropout rate increases from 0.2 to 0.5. For instance, \algo is approximately 10\% better than the baselines at a 0.5 dropout rate in Natural Questions (closed-book), compared to a 2\% improvement at a 0.2 dropout rate. Notably, comparing Figure \ref{subfig:adapfu_dp2} and \ref{subfig:adapfu_dp5}, we notice that \algo is mildly affected by changes in the dropout rate, while the baselines suffer significant degradation as the dropout rate increases. This stability can be attributed to round-by-round emulator alignment, where the non-compressed part of the full model is set as an anchor, regardless of the dropout rate.  Consequently, this approach stabilizes the adapter training process, ensuring that adapters of the same size achieve similar performance across varying dropout rates.

\subsection{Discussion on Computation and Communication Overhead} \label{subsec:ablation}

Table \ref{tab:computation} presents the computation and communication overhead of different methods under different dropout rates. As mentioned in the experimental setting, all algorithms have been applied with LoRA, and therefore, the number of trainable parameters dramatically reduces. From the clients' perspectives, the number of trainable parameters is determined by the number of decoder layers in the adapter. Apparently, \algo (Adapter 2) should be with the minimum number of trainable parameters among other methods. 

The computation costs in Table \ref{tab:computation} are measured by per-token floating point operation (FLOP/token). As we can see, the proposed \algo costs less overhead than offsite-tuning and FedOT. The difference arises on account of the position of the trainable parameters. The adapter of the proposed \algo is near the output layer. As for offsite-tuning and FedOT, the adapters are located separately at the top and the bottom two layers, thereby consuming more computation costs in the backward propagation for transmitting the derivative from the bottom to the top. 

However, our proposed method may require more communication overhead than the baselines. This is because the server should transmit the LoRA parameters of both the adapter and the emulator to the clients in our proposed method, while in offsite-tuning and FedOT, the server merely transmits the aggregated LoRA of the adapter to the clients. However, the overall cost is trivial, compared to the full LLM transmission at a cost of 28GB.  

\begin{table}
\renewcommand{\arraystretch}{1.3}
    \centering
    \caption{Computation and communication costs of different methods under different dropout rates at client side.}
    \vspace{-5px}
    \resizebox{\linewidth}{!}{
    \begin{tabular}{llcccccc}
        \Xhline{1.5pt}
        \makecell[l]{Dropout\\Rate ($\beta$)} & Methods & \makecell[c]{\#Layers in\\Adapter} & \makecell[c]{\#Layers in\\Emulator} &  \makecell[c]{Trainable\\Param. (M)} & \makecell[c]{Comp. Costs\\(GFLOP/token)} & \makecell[c]{Comm. Costs\\(MB/round)}
        \\\hline
        
        \multirow{5}{*}{\LARGE{$\beta=0.2$}} & \makecell[l]{Offsite-tuning\\/FedOT} & \LARGE{4} & \LARGE{22} & \LARGE{0.524} & \LARGE{10.33} & \LARGE{4.19}  \\\cdashline{2-7}
        & \makecell[l]{\algo\\(Adapter 2)} & \LARGE{2} & \LARGE{24} & \LARGE{0.262} & \LARGE{5.47} & \LARGE{14.68}  \\\cdashline{2-7}
        & \makecell[l]{\algo\\(Adapter 4)} & \LARGE{4} & \LARGE{22} & \LARGE{0.524} & \LARGE{5.87} & \LARGE{15.73}  \\ \hline
        
        \multirow{5}{*}{\LARGE{$\beta=0.5$}} & \makecell[l]{Offsite-tuning\\/FedOT} & \LARGE{4} & \LARGE{14} & \LARGE{0.524} & \LARGE{7.09} & \LARGE{4.19}  \\\cdashline{2-7}
        & \makecell[l]{\algo\\(Adapter 2)} & \LARGE{2} & \LARGE{15} & \LARGE{0.262} & \LARGE{3.65} & \LARGE{9.96}  \\\cdashline{2-7}
        & \makecell[l]{\algo\\(Adapter 4)} & \LARGE{4} & \LARGE{14} & \LARGE{0.524} & \LARGE{4.25} & \LARGE{11.53} \\ 
        \Xhline{1.5pt}
    \end{tabular}
    }
    \vspace{-10pt}
    \label{tab:computation}
\end{table}

\section{Conclusion}

In this paper, we introduce \algo, a federated learning algorithm that avoids full model fine-tuning while substantially reducing computation overhead. Specifically, we compress the LLM and divide it into two components, namely, an emulator and an adapter. By formulating a bi-level optimization problem, our proposed \algo ensures that the emulator partially simulates the original LLM, while the adapter focuses on learning domain-specific linguistic patterns. Extensive experiments show the superiority of the proposed \algo working with LLaMA-2, where it can achieve significant accuracy improvement than the existing baselines (i.e., Offsite-tuning and FedOT) in all tasks (i.e., math problem-solving, code generation, and question answering). 

\begin{acks}
The authors would like to thank the anonymous reviewers for their constructive comments. This work is supported in part by the US National Science Foundation under grants NSF-IIS 1747614 and NSF-IIS 2141037. Any opinions, findings, and conclusions or recommendations expressed in this material are those of the author(s) and do not necessarily reflect the views of the National Science Foundation. 
\end{acks}

\bibliographystyle{ACM-Reference-Format}
\balance
\bibliography{sample-base}


\begin{thebibliography}{50}


\ifx \showCODEN    \undefined \def \showCODEN     #1{\unskip}     \fi
\ifx \showDOI      \undefined \def \showDOI       #1{#1}\fi
\ifx \showISBNx    \undefined \def \showISBNx     #1{\unskip}     \fi
\ifx \showISBNxiii \undefined \def \showISBNxiii  #1{\unskip}     \fi
\ifx \showISSN     \undefined \def \showISSN      #1{\unskip}     \fi
\ifx \showLCCN     \undefined \def \showLCCN      #1{\unskip}     \fi
\ifx \shownote     \undefined \def \shownote      #1{#1}          \fi
\ifx \showarticletitle \undefined \def \showarticletitle #1{#1}   \fi
\ifx \showURL      \undefined \def \showURL       {\relax}        \fi
\providecommand\bibfield[2]{#2}
\providecommand\bibinfo[2]{#2}
\providecommand\natexlab[1]{#1}
\providecommand\showeprint[2][]{arXiv:#2}

\bibitem[Acar et~al\mbox{.}(2020)]%
        {acar2020federated}
\bibfield{author}{\bibinfo{person}{Durmus Alp~Emre Acar}, \bibinfo{person}{Yue Zhao}, \bibinfo{person}{Ramon Matas}, \bibinfo{person}{Matthew Mattina}, \bibinfo{person}{Paul Whatmough}, {and} \bibinfo{person}{Venkatesh Saligrama}.} \bibinfo{year}{2020}\natexlab{}.
\newblock \showarticletitle{Federated Learning Based on Dynamic Regularization}. In \bibinfo{booktitle}{\emph{Proc. of International Conference on Learning Representations (ICLR'20)}}.
\newblock


\bibitem[CCPA(2023)]%
        {CCPA}
\bibfield{author}{\bibinfo{person}{CCPA}.} \bibinfo{year}{2023}\natexlab{}.
\newblock \bibinfo{booktitle}{\emph{California Consumer Privacy Act (CCPA)}}.
\newblock
\urldef\tempurl%
\url{https://oag.ca.gov/privacy/ccpa}
\showURL{%
\tempurl}


\bibitem[Chaudhary(2023)]%
        {codealpaca}
\bibfield{author}{\bibinfo{person}{Sahil Chaudhary}.} \bibinfo{year}{2023}\natexlab{}.
\newblock \bibinfo{title}{Code Alpaca: An Instruction-following LLaMA model for code generation}.
\newblock \bibinfo{howpublished}{\url{https://github.com/sahil280114/codealpaca}}.
\newblock


\bibitem[Chen et~al\mbox{.}(2021)]%
        {chen2021evaluating}
\bibfield{author}{\bibinfo{person}{Mark Chen}, \bibinfo{person}{Jerry Tworek}, \bibinfo{person}{Heewoo Jun}, \bibinfo{person}{Qiming Yuan}, \bibinfo{person}{Henrique Ponde de~Oliveira Pinto}, \bibinfo{person}{Jared Kaplan}, \bibinfo{person}{Harri Edwards}, \bibinfo{person}{Yuri Burda}, \bibinfo{person}{Nicholas Joseph}, \bibinfo{person}{Greg Brockman}, {et~al\mbox{.}}} \bibinfo{year}{2021}\natexlab{}.
\newblock \showarticletitle{Evaluating large language models trained on code}.
\newblock \bibinfo{journal}{\emph{arXiv preprint arXiv:2107.03374}} (\bibinfo{year}{2021}).
\newblock


\bibitem[Cobbe et~al\mbox{.}(2021)]%
        {cobbe2021training}
\bibfield{author}{\bibinfo{person}{Karl Cobbe}, \bibinfo{person}{Vineet Kosaraju}, \bibinfo{person}{Mohammad Bavarian}, \bibinfo{person}{Mark Chen}, \bibinfo{person}{Heewoo Jun}, \bibinfo{person}{Lukasz Kaiser}, \bibinfo{person}{Matthias Plappert}, \bibinfo{person}{Jerry Tworek}, \bibinfo{person}{Jacob Hilton}, \bibinfo{person}{Reiichiro Nakano}, {et~al\mbox{.}}} \bibinfo{year}{2021}\natexlab{}.
\newblock \showarticletitle{Training verifiers to solve math word problems}.
\newblock \bibinfo{journal}{\emph{arXiv preprint arXiv:2110.14168}} (\bibinfo{year}{2021}).
\newblock


\bibitem[Conover et~al\mbox{.}(2023)]%
        {DatabricksBlog2023DollyV2}
\bibfield{author}{\bibinfo{person}{Mike Conover}, \bibinfo{person}{Matt Hayes}, \bibinfo{person}{Ankit Mathur}, \bibinfo{person}{Jianwei Xie}, \bibinfo{person}{Jun Wan}, \bibinfo{person}{Sam Shah}, \bibinfo{person}{Ali Ghodsi}, \bibinfo{person}{Patrick Wendell}, \bibinfo{person}{Matei Zaharia}, {and} \bibinfo{person}{Reynold Xin}.} \bibinfo{year}{2023}\natexlab{}.
\newblock \bibinfo{booktitle}{\emph{Free Dolly: Introducing the World's First Truly Open Instruction-Tuned LLM}}.
\newblock
\urldef\tempurl%
\url{https://www.databricks.com/blog/2023/04/12/dolly-first-open-commercially-viable-instruction-tuned-llm}
\showURL{%
\tempurl}


\bibitem[Cui et~al\mbox{.}(2023)]%
        {cui2023chatlaw}
\bibfield{author}{\bibinfo{person}{Jiaxi Cui}, \bibinfo{person}{Zongjian Li}, \bibinfo{person}{Yang Yan}, \bibinfo{person}{Bohua Chen}, {and} \bibinfo{person}{Li Yuan}.} \bibinfo{year}{2023}\natexlab{}.
\newblock \showarticletitle{Chatlaw: Open-source legal large language model with integrated external knowledge bases}.
\newblock \bibinfo{journal}{\emph{arXiv preprint arXiv:2306.16092}} (\bibinfo{year}{2023}).
\newblock


\bibitem[GDPR(2016)]%
        {GDPR2016a}
\bibfield{author}{\bibinfo{person}{GDPR}.} \bibinfo{year}{2016}\natexlab{}.
\newblock \bibinfo{booktitle}{\emph{Regulation ({EU}) 2016/679 of the {European} {Parliament} and of the {Council}}}.
\newblock
\urldef\tempurl%
\url{https://data.europa.eu/eli/reg/2016/679/oj}
\showURL{%
\tempurl}


\bibitem[He et~al\mbox{.}(2023)]%
        {he2023gluefl}
\bibfield{author}{\bibinfo{person}{Shiqi He}, \bibinfo{person}{Qifan Yan}, \bibinfo{person}{Feijie Wu}, \bibinfo{person}{Lanjun Wang}, \bibinfo{person}{Mathias L{\'e}cuyer}, {and} \bibinfo{person}{Ivan Beschastnikh}.} \bibinfo{year}{2023}\natexlab{}.
\newblock \showarticletitle{GlueFL: Reconciling Client Sampling and Model Masking for Bandwidth Efficient Federated Learning}.
\newblock \bibinfo{journal}{\emph{Proc. of Machine Learning and Systems (MLSys'23)}}.
\newblock


\bibitem[Hinton et~al\mbox{.}(2015)]%
        {hinton2015distilling}
\bibfield{author}{\bibinfo{person}{Geoffrey Hinton}, \bibinfo{person}{Oriol Vinyals}, {and} \bibinfo{person}{Jeff Dean}.} \bibinfo{year}{2015}\natexlab{}.
\newblock \showarticletitle{Distilling the knowledge in a neural network}.
\newblock \bibinfo{journal}{\emph{arXiv preprint arXiv:1503.02531}} (\bibinfo{year}{2015}).
\newblock


\bibitem[Hu et~al\mbox{.}(2021)]%
        {hu2021lora}
\bibfield{author}{\bibinfo{person}{Edward~J Hu}, \bibinfo{person}{Phillip Wallis}, \bibinfo{person}{Zeyuan Allen-Zhu}, \bibinfo{person}{Yuanzhi Li}, \bibinfo{person}{Shean Wang}, \bibinfo{person}{Lu Wang}, \bibinfo{person}{Weizhu Chen}, {et~al\mbox{.}}} \bibinfo{year}{2021}\natexlab{}.
\newblock \showarticletitle{LoRA: Low-Rank Adaptation of Large Language Models}. In \bibinfo{booktitle}{\emph{Proc. of International Conference on Learning Representations (ICLR'21)}}.
\newblock


\bibitem[Karimireddy et~al\mbox{.}(2020)]%
        {karimireddy2020scaffold}
\bibfield{author}{\bibinfo{person}{Sai~Praneeth Karimireddy}, \bibinfo{person}{Satyen Kale}, \bibinfo{person}{Mehryar Mohri}, \bibinfo{person}{Sashank Reddi}, \bibinfo{person}{Sebastian Stich}, {and} \bibinfo{person}{Ananda~Theertha Suresh}.} \bibinfo{year}{2020}\natexlab{}.
\newblock \showarticletitle{Scaffold: Stochastic controlled averaging for federated learning}. In \bibinfo{booktitle}{\emph{Proc. of International conference on machine learning (ICML'20)}}. \bibinfo{pages}{5132--5143}.
\newblock


\bibitem[Kuang et~al\mbox{.}(2024)]%
        {kuang2023federatedscope}
\bibfield{author}{\bibinfo{person}{Weirui Kuang}, \bibinfo{person}{Bingchen Qian}, \bibinfo{person}{Zitao Li}, \bibinfo{person}{Daoyuan Chen}, \bibinfo{person}{Dawei Gao}, \bibinfo{person}{Xuchen Pan}, \bibinfo{person}{Yuexiang Xie}, \bibinfo{person}{Yaliang Li}, \bibinfo{person}{Bolin Ding}, {and} \bibinfo{person}{Jingren Zhou}.} \bibinfo{year}{2024}\natexlab{}.
\newblock \showarticletitle{FederatedScope-LLM: A Comprehensive Package for Fine-tuning Large Language Models in Federated Learning}. In \bibinfo{booktitle}{\emph{Proc. of the ACM SIGKDD Conference on Knowledge Discovery and Data Mining (KDD'24)}}.
\newblock


\bibitem[Lester et~al\mbox{.}(2021)]%
        {lester2021power}
\bibfield{author}{\bibinfo{person}{Brian Lester}, \bibinfo{person}{Rami Al-Rfou}, {and} \bibinfo{person}{Noah Constant}.} \bibinfo{year}{2021}\natexlab{}.
\newblock \showarticletitle{The Power of Scale for Parameter-Efficient Prompt Tuning}. In \bibinfo{booktitle}{\emph{Proc. of the Conference on Empirical Methods in Natural Language Processing (EMNLP'21)}}. \bibinfo{pages}{3045--3059}.
\newblock


\bibitem[Li et~al\mbox{.}(2019)]%
        {li2019convergence}
\bibfield{author}{\bibinfo{person}{Xiang Li}, \bibinfo{person}{Kaixuan Huang}, \bibinfo{person}{Wenhao Yang}, \bibinfo{person}{Shusen Wang}, {and} \bibinfo{person}{Zhihua Zhang}.} \bibinfo{year}{2019}\natexlab{}.
\newblock \showarticletitle{On the Convergence of FedAvg on Non-IID Data}. In \bibinfo{booktitle}{\emph{Proc. of International Conference on Learning Representations (ICLR'19)}}.
\newblock


\bibitem[Li and Liang(2021)]%
        {li2021prefix}
\bibfield{author}{\bibinfo{person}{Xiang~Lisa Li} {and} \bibinfo{person}{Percy Liang}.} \bibinfo{year}{2021}\natexlab{}.
\newblock \showarticletitle{Prefix-Tuning: Optimizing Continuous Prompts for Generation}. In \bibinfo{booktitle}{\emph{Proc. of the Annual Meeting of the Association for Computational Linguistics and the International Joint Conference on Natural Language Processing (ACL/IJNLP'21)}}. \bibinfo{pages}{4582--4597}.
\newblock


\bibitem[Liang et~al\mbox{.}(2022)]%
        {liang2022holistic}
\bibfield{author}{\bibinfo{person}{Percy Liang}, \bibinfo{person}{Rishi Bommasani}, \bibinfo{person}{Tony Lee}, \bibinfo{person}{Dimitris Tsipras}, \bibinfo{person}{Dilara Soylu}, \bibinfo{person}{Michihiro Yasunaga}, \bibinfo{person}{Yian Zhang}, \bibinfo{person}{Deepak Narayanan}, \bibinfo{person}{Yuhuai Wu}, \bibinfo{person}{Ananya Kumar}, {et~al\mbox{.}}} \bibinfo{year}{2022}\natexlab{}.
\newblock \showarticletitle{Holistic evaluation of language models}.
\newblock \bibinfo{journal}{\emph{arXiv preprint arXiv:2211.09110}} (\bibinfo{year}{2022}).
\newblock


\bibitem[Lin et~al\mbox{.}(2020)]%
        {lin2020ensemble}
\bibfield{author}{\bibinfo{person}{Tao Lin}, \bibinfo{person}{Lingjing Kong}, \bibinfo{person}{Sebastian~U Stich}, {and} \bibinfo{person}{Martin Jaggi}.} \bibinfo{year}{2020}\natexlab{}.
\newblock \showarticletitle{Ensemble distillation for robust model fusion in federated learning}. In \bibinfo{booktitle}{\emph{Proc. of Advances in Neural Information Processing Systems (NeurIPS'20)}}. \bibinfo{pages}{2351--2363}.
\newblock


\bibitem[Lin et~al\mbox{.}(2023)]%
        {lin2023efficient}
\bibfield{author}{\bibinfo{person}{Zihao Lin}, \bibinfo{person}{Yan Sun}, \bibinfo{person}{Yifan Shi}, \bibinfo{person}{Xueqian Wang}, \bibinfo{person}{Lifu Huang}, \bibinfo{person}{Li Shen}, {and} \bibinfo{person}{Dacheng Tao}.} \bibinfo{year}{2023}\natexlab{}.
\newblock \showarticletitle{Efficient federated prompt tuning for black-box large pre-trained models}.
\newblock \bibinfo{journal}{\emph{arXiv preprint arXiv:2310.03123}} (\bibinfo{year}{2023}).
\newblock


\bibitem[Liu et~al\mbox{.}(2023)]%
        {liu2023gpt}
\bibfield{author}{\bibinfo{person}{Xiao Liu}, \bibinfo{person}{Yanan Zheng}, \bibinfo{person}{Zhengxiao Du}, \bibinfo{person}{Ming Ding}, \bibinfo{person}{Yujie Qian}, \bibinfo{person}{Zhilin Yang}, {and} \bibinfo{person}{Jie Tang}.} \bibinfo{year}{2023}\natexlab{}.
\newblock \showarticletitle{GPT understands, too}.
\newblock \bibinfo{journal}{\emph{AI Open}} (\bibinfo{year}{2023}).
\newblock


\bibitem[Loshchilov and Hutter(2018)]%
        {loshchilov2017decoupled}
\bibfield{author}{\bibinfo{person}{Ilya Loshchilov} {and} \bibinfo{person}{Frank Hutter}.} \bibinfo{year}{2018}\natexlab{}.
\newblock \showarticletitle{Decoupled Weight Decay Regularization}. In \bibinfo{booktitle}{\emph{Proc. of International Conference on Learning Representations (ICLR'18)}}.
\newblock


\bibitem[McMahan et~al\mbox{.}(2017)]%
        {mcmahan2017communication}
\bibfield{author}{\bibinfo{person}{Brendan McMahan}, \bibinfo{person}{Eider Moore}, \bibinfo{person}{Daniel Ramage}, \bibinfo{person}{Seth Hampson}, {and} \bibinfo{person}{Blaise~Aguera y Arcas}.} \bibinfo{year}{2017}\natexlab{}.
\newblock \showarticletitle{Communication-efficient learning of deep networks from decentralized data}. In \bibinfo{booktitle}{\emph{Proc. of Artificial intelligence and statistics (AISTAT'17)}}. \bibinfo{pages}{1273--1282}.
\newblock


\bibitem[Nay et~al\mbox{.}(2024)]%
        {nay2023large}
\bibfield{author}{\bibinfo{person}{John~J Nay}, \bibinfo{person}{David Karamardian}, \bibinfo{person}{Sarah~B Lawsky}, \bibinfo{person}{Wenting Tao}, \bibinfo{person}{Meghana Bhat}, \bibinfo{person}{Raghav Jain}, \bibinfo{person}{Aaron~Travis Lee}, \bibinfo{person}{Jonathan~H Choi}, {and} \bibinfo{person}{Jungo Kasai}.} \bibinfo{year}{2024}\natexlab{}.
\newblock \showarticletitle{Large language models as tax attorneys: a case study in legal capabilities emergence}.
\newblock \bibinfo{journal}{\emph{Philosophical Transactions of the Royal Society A}} \bibinfo{volume}{382}, \bibinfo{number}{2270} (\bibinfo{year}{2024}), \bibinfo{pages}{20230159}.
\newblock


\bibitem[OpenAI(2023)]%
        {chatgptfinetune}
\bibfield{author}{\bibinfo{person}{OpenAI}.} \bibinfo{year}{2023}\natexlab{}.
\newblock \bibinfo{title}{Fine-tuning - OpenAI API}.
\newblock \bibinfo{howpublished}{\url{https://platform.openai.com/docs/guides/fine-tuning}}.
\newblock
\newblock
\shownote{Accessed: 2023-09-29}.


\bibitem[Ouyang et~al\mbox{.}(2022)]%
        {ouyang2022training}
\bibfield{author}{\bibinfo{person}{Long Ouyang}, \bibinfo{person}{Jeffrey Wu}, \bibinfo{person}{Xu Jiang}, \bibinfo{person}{Diogo Almeida}, \bibinfo{person}{Carroll Wainwright}, \bibinfo{person}{Pamela Mishkin}, \bibinfo{person}{Chong Zhang}, \bibinfo{person}{Sandhini Agarwal}, \bibinfo{person}{Katarina Slama}, \bibinfo{person}{Alex Ray}, {et~al\mbox{.}}} \bibinfo{year}{2022}\natexlab{}.
\newblock \showarticletitle{Training language models to follow instructions with human feedback}. In \bibinfo{booktitle}{\emph{Proc. of Advances in Neural Information Processing Systems (NeurIPS'22)}}. \bibinfo{pages}{27730--27744}.
\newblock


\bibitem[Sajjad et~al\mbox{.}(2023)]%
        {sajjad2023effect}
\bibfield{author}{\bibinfo{person}{Hassan Sajjad}, \bibinfo{person}{Fahim Dalvi}, \bibinfo{person}{Nadir Durrani}, {and} \bibinfo{person}{Preslav Nakov}.} \bibinfo{year}{2023}\natexlab{}.
\newblock \showarticletitle{On the effect of dropping layers of pre-trained transformer models}.
\newblock \bibinfo{journal}{\emph{Computer Speech \& Language}}  \bibinfo{volume}{77} (\bibinfo{year}{2023}), \bibinfo{pages}{101429}.
\newblock


\bibitem[Singhal et~al\mbox{.}(2023)]%
        {singhal2023large}
\bibfield{author}{\bibinfo{person}{Karan Singhal}, \bibinfo{person}{Shekoofeh Azizi}, \bibinfo{person}{Tao Tu}, \bibinfo{person}{S~Sara Mahdavi}, \bibinfo{person}{Jason Wei}, \bibinfo{person}{Hyung~Won Chung}, \bibinfo{person}{Nathan Scales}, \bibinfo{person}{Ajay Tanwani}, \bibinfo{person}{Heather Cole-Lewis}, \bibinfo{person}{Stephen Pfohl}, {et~al\mbox{.}}} \bibinfo{year}{2023}\natexlab{}.
\newblock \showarticletitle{Large language models encode clinical knowledge}.
\newblock \bibinfo{journal}{\emph{Nature}} \bibinfo{volume}{620}, \bibinfo{number}{7972} (\bibinfo{year}{2023}), \bibinfo{pages}{172--180}.
\newblock


\bibitem[Sordoni et~al\mbox{.}(2023)]%
        {sordoni2023joint}
\bibfield{author}{\bibinfo{person}{Alessandro Sordoni}, \bibinfo{person}{Xingdi Yuan}, \bibinfo{person}{Marc-Alexandre C{\^o}t{\'e}}, \bibinfo{person}{Matheus Pereira}, \bibinfo{person}{Adam Trischler}, \bibinfo{person}{Ziang Xiao}, \bibinfo{person}{Arian Hosseini}, \bibinfo{person}{Friederike Niedtner}, {and} \bibinfo{person}{Nicolas Le~Roux}.} \bibinfo{year}{2023}\natexlab{}.
\newblock \showarticletitle{Joint prompt optimization of stacked llms using variational inference}. In \bibinfo{booktitle}{\emph{Proc. of Advances in Neural Information Processing Systems (NeurIPS'23)}}.
\newblock


\bibitem[Sun et~al\mbox{.}(2023)]%
        {sun2023fedbpt}
\bibfield{author}{\bibinfo{person}{Jingwei Sun}, \bibinfo{person}{Ziyue Xu}, \bibinfo{person}{Hongxu Yin}, \bibinfo{person}{Dong Yang}, \bibinfo{person}{Daguang Xu}, \bibinfo{person}{Yiran Chen}, {and} \bibinfo{person}{Holger~R Roth}.} \bibinfo{year}{2023}\natexlab{}.
\newblock \showarticletitle{FedBPT: Efficient Federated Black-box Prompt Tuning for Large Language Models}.
\newblock \bibinfo{journal}{\emph{arXiv preprint arXiv:2310.01467}} (\bibinfo{year}{2023}).
\newblock


\bibitem[Sun et~al\mbox{.}(2024)]%
        {sun2024improving}
\bibfield{author}{\bibinfo{person}{Youbang Sun}, \bibinfo{person}{Zitao Li}, \bibinfo{person}{Yaliang Li}, {and} \bibinfo{person}{Bolin Ding}.} \bibinfo{year}{2024}\natexlab{}.
\newblock \showarticletitle{Improving Lo{RA} in Privacy-preserving Federated Learning}. In \bibinfo{booktitle}{\emph{Proc. of The International Conference on Learning Representations (ICLR'24)}}.
\newblock


\bibitem[Taori et~al\mbox{.}(2023)]%
        {alpaca}
\bibfield{author}{\bibinfo{person}{Rohan Taori}, \bibinfo{person}{Ishaan Gulrajani}, \bibinfo{person}{Tianyi Zhang}, \bibinfo{person}{Yann Dubois}, \bibinfo{person}{Xuechen Li}, \bibinfo{person}{Carlos Guestrin}, \bibinfo{person}{Percy Liang}, {and} \bibinfo{person}{Tatsunori~B. Hashimoto}.} \bibinfo{year}{2023}\natexlab{}.
\newblock \bibinfo{title}{Stanford Alpaca: An Instruction-following LLaMA model}.
\newblock \bibinfo{howpublished}{\url{https://github.com/tatsu-lab/stanford_alpaca}}.
\newblock


\bibitem[Thirunavukarasu et~al\mbox{.}(2023)]%
        {thirunavukarasu2023large}
\bibfield{author}{\bibinfo{person}{Arun~James Thirunavukarasu}, \bibinfo{person}{Darren Shu~Jeng Ting}, \bibinfo{person}{Kabilan Elangovan}, \bibinfo{person}{Laura Gutierrez}, \bibinfo{person}{Ting~Fang Tan}, {and} \bibinfo{person}{Daniel Shu~Wei Ting}.} \bibinfo{year}{2023}\natexlab{}.
\newblock \showarticletitle{Large language models in medicine}.
\newblock \bibinfo{journal}{\emph{Nature medicine}} \bibinfo{volume}{29}, \bibinfo{number}{8} (\bibinfo{year}{2023}), \bibinfo{pages}{1930--1940}.
\newblock


\bibitem[Touvron et~al\mbox{.}(2023a)]%
        {touvron2302llama}
\bibfield{author}{\bibinfo{person}{Hugo Touvron}, \bibinfo{person}{Thibaut Lavril}, \bibinfo{person}{Gautier Izacard}, \bibinfo{person}{Xavier Martinet}, \bibinfo{person}{Marie-Anne Lachaux}, \bibinfo{person}{Timoth{\'e}e Lacroix}, \bibinfo{person}{Baptiste Rozi{\`e}re}, \bibinfo{person}{Naman Goyal}, \bibinfo{person}{Eric Hambro}, \bibinfo{person}{Faisal Azhar}, {et~al\mbox{.}}} \bibinfo{year}{2023}\natexlab{a}.
\newblock \showarticletitle{Llama: Open and efficient foundation language models}.
\newblock \bibinfo{journal}{\emph{arXiv preprint arXiv:2302.13971}} (\bibinfo{year}{2023}).
\newblock


\bibitem[Touvron et~al\mbox{.}(2023b)]%
        {touvron2023llama}
\bibfield{author}{\bibinfo{person}{Hugo Touvron}, \bibinfo{person}{Louis Martin}, \bibinfo{person}{Kevin Stone}, \bibinfo{person}{Peter Albert}, \bibinfo{person}{Amjad Almahairi}, \bibinfo{person}{Yasmine Babaei}, \bibinfo{person}{Nikolay Bashlykov}, \bibinfo{person}{Soumya Batra}, \bibinfo{person}{Prajjwal Bhargava}, \bibinfo{person}{Shruti Bhosale}, {et~al\mbox{.}}} \bibinfo{year}{2023}\natexlab{b}.
\newblock \showarticletitle{Llama 2: Open foundation and fine-tuned chat models}.
\newblock \bibinfo{journal}{\emph{arXiv preprint arXiv:2307.09288}} (\bibinfo{year}{2023}).
\newblock


\bibitem[Wang et~al\mbox{.}(2023a)]%
        {wang2023dafkd}
\bibfield{author}{\bibinfo{person}{Haozhao Wang}, \bibinfo{person}{Yichen Li}, \bibinfo{person}{Wenchao Xu}, \bibinfo{person}{Ruixuan Li}, \bibinfo{person}{Yufeng Zhan}, {and} \bibinfo{person}{Zhigang Zeng}.} \bibinfo{year}{2023}\natexlab{a}.
\newblock \showarticletitle{Dafkd: Domain-aware federated knowledge distillation}. In \bibinfo{booktitle}{\emph{Proc. of the IEEE/CVF conference on Computer Vision and Pattern Recognition (CVPR'23)}}. \bibinfo{pages}{20412--20421}.
\newblock


\bibitem[Wang et~al\mbox{.}(2023b)]%
        {wang2023fedcda}
\bibfield{author}{\bibinfo{person}{Haozhao Wang}, \bibinfo{person}{Haoran Xu}, \bibinfo{person}{Yichen Li}, \bibinfo{person}{Yuan Xu}, \bibinfo{person}{Ruixuan Li}, {and} \bibinfo{person}{Tianwei Zhang}.} \bibinfo{year}{2023}\natexlab{b}.
\newblock \showarticletitle{FedCDA: Federated Learning with Cross-rounds Divergence-aware Aggregation}. In \bibinfo{booktitle}{\emph{Proc. of The International Conference on Learning Representations (ICLR'23)}}.
\newblock


\bibitem[Wang et~al\mbox{.}(2022)]%
        {wang2022fedkc}
\bibfield{author}{\bibinfo{person}{Haoyu Wang}, \bibinfo{person}{Handong Zhao}, \bibinfo{person}{Yaqing Wang}, \bibinfo{person}{Tong Yu}, \bibinfo{person}{Jiuxiang Gu}, {and} \bibinfo{person}{Jing Gao}.} \bibinfo{year}{2022}\natexlab{}.
\newblock \showarticletitle{FedKC: Federated knowledge composition for multilingual natural language understanding}. In \bibinfo{booktitle}{\emph{Proc. of the ACM Web Conference 2022 (WWW'22)}}. \bibinfo{pages}{1839--1850}.
\newblock


\bibitem[Wang et~al\mbox{.}(2020)]%
        {wang2020tackling}
\bibfield{author}{\bibinfo{person}{Jianyu Wang}, \bibinfo{person}{Qinghua Liu}, \bibinfo{person}{Hao Liang}, \bibinfo{person}{Gauri Joshi}, {and} \bibinfo{person}{H~Vincent Poor}.} \bibinfo{year}{2020}\natexlab{}.
\newblock \showarticletitle{Tackling the objective inconsistency problem in heterogeneous federated optimization}. In \bibinfo{booktitle}{\emph{Proc. of Advances in neural information processing systems (NeurIPS'20)}}. \bibinfo{pages}{7611--7623}.
\newblock


\bibitem[Wang et~al\mbox{.}(2023c)]%
        {wang2023chatcad}
\bibfield{author}{\bibinfo{person}{Sheng Wang}, \bibinfo{person}{Zihao Zhao}, \bibinfo{person}{Xi Ouyang}, \bibinfo{person}{Qian Wang}, {and} \bibinfo{person}{Dinggang Shen}.} \bibinfo{year}{2023}\natexlab{c}.
\newblock \showarticletitle{Chatcad: Interactive computer-aided diagnosis on medical image using large language models}.
\newblock \bibinfo{journal}{\emph{arXiv preprint arXiv:2302.07257}} (\bibinfo{year}{2023}).
\newblock


\bibitem[Wei et~al\mbox{.}(2021)]%
        {wei2021finetuned}
\bibfield{author}{\bibinfo{person}{Jason Wei}, \bibinfo{person}{Maarten Bosma}, \bibinfo{person}{Vincent Zhao}, \bibinfo{person}{Kelvin Guu}, \bibinfo{person}{Adams~Wei Yu}, \bibinfo{person}{Brian Lester}, \bibinfo{person}{Nan Du}, \bibinfo{person}{Andrew~M Dai}, {and} \bibinfo{person}{Quoc~V Le}.} \bibinfo{year}{2021}\natexlab{}.
\newblock \showarticletitle{Finetuned Language Models are Zero-Shot Learners}. In \bibinfo{booktitle}{\emph{Proc. of International Conference on Learning Representations (ICLR'21)}}.
\newblock


\bibitem[Wei et~al\mbox{.}(2022)]%
        {wei2022chain}
\bibfield{author}{\bibinfo{person}{Jason Wei}, \bibinfo{person}{Xuezhi Wang}, \bibinfo{person}{Dale Schuurmans}, \bibinfo{person}{Maarten Bosma}, \bibinfo{person}{Fei Xia}, \bibinfo{person}{Ed Chi}, \bibinfo{person}{Quoc~V Le}, \bibinfo{person}{Denny Zhou}, {et~al\mbox{.}}} \bibinfo{year}{2022}\natexlab{}.
\newblock \showarticletitle{Chain-of-thought prompting elicits reasoning in large language models}. In \bibinfo{booktitle}{\emph{Proc. of Advances in Neural Information Processing Systems (NeurIPS'22)}}. \bibinfo{pages}{24824--24837}.
\newblock


\bibitem[Wu et~al\mbox{.}(2023)]%
        {wu2023anchor}
\bibfield{author}{\bibinfo{person}{Feijie Wu}, \bibinfo{person}{Song Guo}, \bibinfo{person}{Zhihao Qu}, \bibinfo{person}{Shiqi He}, \bibinfo{person}{Ziming Liu}, {and} \bibinfo{person}{Jing Gao}.} \bibinfo{year}{2023}\natexlab{}.
\newblock \showarticletitle{Anchor sampling for federated learning with partial client participation}. In \bibinfo{booktitle}{\emph{Proc. of International Conference on Machine Learning (ICML'23)}}. \bibinfo{pages}{37379--37416}.
\newblock


\bibitem[Xiao et~al\mbox{.}(2023)]%
        {xiao2023offsite}
\bibfield{author}{\bibinfo{person}{Guangxuan Xiao}, \bibinfo{person}{Ji Lin}, {and} \bibinfo{person}{Song Han}.} \bibinfo{year}{2023}\natexlab{}.
\newblock \showarticletitle{Offsite-tuning: Transfer learning without full model}.
\newblock \bibinfo{journal}{\emph{arXiv preprint arXiv:2302.04870}} (\bibinfo{year}{2023}).
\newblock


\bibitem[Xie et~al\mbox{.}(2023)]%
        {xie2022federatedscope}
\bibfield{author}{\bibinfo{person}{Yuexiang Xie}, \bibinfo{person}{Zhen Wang}, \bibinfo{person}{Dawei Gao}, \bibinfo{person}{Daoyuan Chen}, \bibinfo{person}{Liuyi Yao}, \bibinfo{person}{Weirui Kuang}, \bibinfo{person}{Yaliang Li}, \bibinfo{person}{Bolin Ding}, {and} \bibinfo{person}{Jingren Zhou}.} \bibinfo{year}{2023}\natexlab{}.
\newblock \showarticletitle{FederatedScope: A Flexible Federated Learning Platform for Heterogeneity}. In \bibinfo{booktitle}{\emph{Proc. of the VLDB Endowment (VLDB'23)}}. \bibinfo{pages}{1059--1072}.
\newblock


\bibitem[Yi et~al\mbox{.}(2023)]%
        {yi2023fedlora}
\bibfield{author}{\bibinfo{person}{Liping Yi}, \bibinfo{person}{Han Yu}, \bibinfo{person}{Gang Wang}, {and} \bibinfo{person}{Xiaoguang Liu}.} \bibinfo{year}{2023}\natexlab{}.
\newblock \showarticletitle{Fedlora: Model-heterogeneous personalized federated learning with lora tuning}.
\newblock \bibinfo{journal}{\emph{arXiv preprint arXiv:2310.13283}} (\bibinfo{year}{2023}).
\newblock


\bibitem[Yosinski et~al\mbox{.}(2014)]%
        {yosinski2014transferable}
\bibfield{author}{\bibinfo{person}{Jason Yosinski}, \bibinfo{person}{Jeff Clune}, \bibinfo{person}{Yoshua Bengio}, {and} \bibinfo{person}{Hod Lipson}.} \bibinfo{year}{2014}\natexlab{}.
\newblock \showarticletitle{How transferable are features in deep neural networks?}. In \bibinfo{booktitle}{\emph{Proc. of Advances in neural information processing systems (NeurIPS'14)}}.
\newblock


\bibitem[Zhang et~al\mbox{.}(2021)]%
        {zhang2021parameterized}
\bibfield{author}{\bibinfo{person}{Jie Zhang}, \bibinfo{person}{Song Guo}, \bibinfo{person}{Xiaosong Ma}, \bibinfo{person}{Haozhao Wang}, \bibinfo{person}{Wenchao Xu}, {and} \bibinfo{person}{Feijie Wu}.} \bibinfo{year}{2021}\natexlab{}.
\newblock \showarticletitle{Parameterized knowledge transfer for personalized federated learning}. In \bibinfo{booktitle}{\emph{Proc. of Advances in Neural Information Processing Systems (NeurIPS'21)}}. \bibinfo{pages}{10092--10104}.
\newblock


\bibitem[Zhang et~al\mbox{.}(2024)]%
        {zhang2023towards}
\bibfield{author}{\bibinfo{person}{Jianyi Zhang}, \bibinfo{person}{Saeed Vahidian}, \bibinfo{person}{Martin Kuo}, \bibinfo{person}{Chunyuan Li}, \bibinfo{person}{Ruiyi Zhang}, \bibinfo{person}{Tong Yu}, \bibinfo{person}{Guoyin Wang}, {and} \bibinfo{person}{Yiran Chen}.} \bibinfo{year}{2024}\natexlab{}.
\newblock \showarticletitle{Towards building the federatedGPT: Federated instruction tuning}. In \bibinfo{booktitle}{\emph{Proc. of IEEE International Conference on Acoustics, Speech and Signal Processing (ICASSP'24)}}. \bibinfo{pages}{6915--6919}.
\newblock


\bibitem[Zhang et~al\mbox{.}(2023)]%
        {zhang2023fedpetuning}
\bibfield{author}{\bibinfo{person}{Zhuo Zhang}, \bibinfo{person}{Yuanhang Yang}, \bibinfo{person}{Yong Dai}, \bibinfo{person}{Qifan Wang}, \bibinfo{person}{Yue Yu}, \bibinfo{person}{Lizhen Qu}, {and} \bibinfo{person}{Zenglin Xu}.} \bibinfo{year}{2023}\natexlab{}.
\newblock \showarticletitle{FedPETuning: When federated learning meets the parameter-efficient tuning methods of pre-trained language models}. In \bibinfo{booktitle}{\emph{Proc. of Annual Meeting of the Association of Computational Linguistics (ACL'23)}}. \bibinfo{pages}{9963--9977}.
\newblock


\bibitem[Zheng et~al\mbox{.}(2023)]%
        {zheng2023codegeex}
\bibfield{author}{\bibinfo{person}{Qinkai Zheng}, \bibinfo{person}{Xiao Xia}, \bibinfo{person}{Xu Zou}, \bibinfo{person}{Yuxiao Dong}, \bibinfo{person}{Shan Wang}, \bibinfo{person}{Yufei Xue}, \bibinfo{person}{Lei Shen}, \bibinfo{person}{Zihan Wang}, \bibinfo{person}{Andi Wang}, \bibinfo{person}{Yang Li}, {et~al\mbox{.}}} \bibinfo{year}{2023}\natexlab{}.
\newblock \showarticletitle{Codegeex: A pre-trained model for code generation with multilingual benchmarking on humaneval-x}. In \bibinfo{booktitle}{\emph{Proc. of the ACM SIGKDD Conference on Knowledge Discovery and Data Mining (KDD'23)}}. \bibinfo{pages}{5673--5684}.
\newblock


\end{thebibliography}

\appendix

\section{Testing Dataset and Evaluation} \label{apdx:eval}

As described in Table \ref{tab:datasets}, we utilize three datasets to assess the fine-tuning performance. In this section, we briefly introduce all these datasets and provide the details about how they evaluate a given LLM. 

\mypara{GSM-8K.} 
We use the GSM-8K test set \cite{cobbe2021training} to evaluate the ability of a large language model (LLM) to solve math problems. This dataset includes "questions" and "ground truth" answers. We assess correctness by determining how often the LLM answers a given question correctly. Following chain of thought (CoT) \cite{wei2022chain}, we prepare a set of sample questions (a.k.a. few-shot prompting) and prompt the LLM to generate step-by-step solutions, ensuring the answers are formatted correctly. Finally, we extract the answers from these solutions and compare them with the ground truth to calculate the correctness rate. 

\mypara{HumanevalX.} This is a task for code autofill, which consists of 164 test samples for five programming languages \cite{zheng2023codegeex}. It is worth noting that we use four of them (i.e., C++, GO, Java, and Python) because there are no JavaScript codes in the training dataset. Each test sample is constituted with ``task id'', ``prompt'' (i.e., Task description with partial codes), ``entry point'' (i.e., the function to be achieved), ``canonical solution'' (i.e., a sampled solution), and ``test'' (i.e., evaluate if the generated code can obtain the correct answer based on the given input). In this task, we use ``prompt'' as the input and generate five versions of codes using a given model. We compile the code and check if it can pass the given ``test''. Let $c$ be the number of correct codes generated by LLM and passed unit tests, and therefore, Pass@k can be computed via 
\begin{equation*}
    \text{Pass@k} = \mathbb{E}_{\text{problems}} \left[1 - \frac{\binom{n-c}{k}}{\binom{n}{k}}\right]
\end{equation*}

\mypara{HELM.} HELM \cite{liang2022holistic} is a benchmark that contains a wide range of NLP tasks. We upload the well-trained models to the benchmark and evaluate them on question-answering tasks, which includes eight datasets, i.e., MMLU, BoolQ, NarrativeQA, Natural Questions (closed-book), Natural Questions (open-book), QuAC, HellaSwag, OpenbookQA. For different tasks, the results come from different metrics, i.e., \textbf{exact match} for \textit{HellaSwag}, \textit{OpenbookQA}, and \textit{MMLU}; \textbf{quasi-exact match} for \textit{BoolQ}; \textbf{F1} for the rest. 



\end{document}